%% file: main.tex
\documentclass[acmtog]{acmart}

\usepackage{booktabs} 

\usepackage{amsmath}
\usepackage{algorithm}
\usepackage{algpseudocode}
\usepackage{graphicx}
\makeatletter
\renewcommand{\ALG@name}{\small ALGORITHM}
\makeatother
\usepackage{caption}
\usepackage{xspace}
\algrenewcommand{\algorithmicindent}{1.1em}
\usepackage{cleveref}
\usepackage{dsfont}
\definecolor{goodindcolor}{HTML}{B9F2F1}
\definecolor{corrupindcolor}{HTML}{DB9EA1}
\input{math_commands}

\usepackage[normalem]{ulem}
\usepackage{listings}
\usepackage{titletoc}
\usepackage[table]{xcolor}

\definecolor{codegreen}{rgb}{0,0.6,0}
\definecolor{codegray}{rgb}{0.5,0.5,0.5}
\definecolor{codepurple}{rgb}{0.58,0,0.82}
\definecolor{backcolour}{rgb}{0.95,0.95,0.92}

\lstdefinestyle{mystyle}{
    backgroundcolor=\color{backcolour},   
    commentstyle=\color{codegreen},
    keywordstyle=\color{magenta},
    numberstyle=\tiny\color{codegray},
    stringstyle=\color{codepurple},
    basicstyle=\ttfamily\footnotesize,
    breakatwhitespace=false,         
    breaklines=true,                 
    captionpos=b,                    
    keepspaces=true,                 
    numbers=left,                    
    numbersep=5pt,                  
    showspaces=false,                
    showstringspaces=false,
    showtabs=false,                  
    tabsize=2
}

\setcopyright{acmcopyright}
\setcopyright{acmlicensed}
\setcopyright{rightsretained}

\copyrightyear{2025}
\acmYear{2025}
\acmConference[SA Conference Papers '25]{SIGGRAPH Asia 2025 Conference Papers}{December 15--18, 2025}{Hong Kong, Hong Kong}
\acmBooktitle{SIGGRAPH Asia 2025 Conference Papers (SA Conference Papers '25), December 15--18, 2025, Hong Kong, Hong Kong}
\acmDOI{10.1145/3757377.3763817}
\acmISBN{979-8-4007-2137-3/2025/12}

\begin{document}

\title{StableMotion: Training Motion Cleanup Models with Unpaired Corrupted Data}

\author{Yuxuan Mu}
\email{yma101@sfu.ca}
\affiliation{%
  \institution{Simon Fraser University}
  \country{Canada}
}

\author{Hung Yu Ling}
\email{hling@ea.com}
\affiliation{%
  \institution{Electronic Arts}
  \country{Canada}
}

\author{Yi Shi}
\email{ysa273@sfu.ca}
\affiliation{%
  \institution{Simon Fraser University}
  \country{Canada}
}

\author{Ismael Baira Ojeda}
\email{ibaira@ea.com}
\affiliation{%
  \institution{Electronic Arts}
  \country{Canada}
}

\author{Pengcheng Xi}
\email{Pengcheng.Xi@nrc-cnrc.gc.ca}
\affiliation{%
  \institution{National Research Council Canada}
  \country{Canada}
}

\author{Chang Shu}
\email{Chang.Shu@nrc-cnrc.gc.ca}
\affiliation{%
  \institution{National Research Council Canada}
  \country{Canada}
}

\author{Fabio Zinno}
\email{fzinno@ea.com}
\affiliation{%
  \institution{Electronic Arts}
  \country{Canada}
}

\author{Xue Bin Peng}
\email{xbpeng@sfu.ca}
\affiliation{%
  \institution{Simon Fraser University}
  \country{Canada}
}
\affiliation{%
  \institution{Nvidia}
  \country{Canada}
}

\newcommand{\et}[2]{${#1}^{\pm{#2}}$}
\newcommand{\etb}[2]{$\mathbf{{#1}}^{\pm{#2}}$}
\newcommand{\ets}[2]{$\underline{{#1}}^{\pm{#2}}$}
\newcommand{\modelterm}[0]{generate-discriminate model}
\newcommand{\yxmu}[1]{\textcolor{blue}{#1}}
\newcommand{\mydel}[1]{\textcolor{red}{\sout{#1}}}
\newcommand{\myadd}[1]{\textcolor{blue}{#1}}
\setlength{\leftmargini}{2em}

\def\soccermocap{SoccerMocap\xspace}
\def\model{StableMotion\xspace}
\def\method{GDM\xspace}
\def\qualvar{QualVar\xspace} 


\begin{abstract}
    Motion capture (mocap) data often exhibits visually jarring artifacts due to inaccurate sensors and post-processing. Cleaning this corrupted data can require substantial manual effort from human experts, which can be a costly and time-consuming process. Previous data-driven motion cleanup methods offer the promise of automating this cleanup process, but often require in-domain paired corrupted-to-clean training data. Constructing such paired datasets requires access to high-quality, relatively artifact-free motion clips, which often necessitates laborious manual cleanup. In this work, we present StableMotion, a simple yet effective method for training motion cleanup models directly from unpaired corrupted datasets that need cleanup. The core component of our method is the introduction of motion quality indicators, which can be easily annotated— through manual labeling or heuristic algorithms—and enable training of quality-aware motion generation models on raw motion data with \emph{mixed} quality. At test time, the model can be prompted to generate high-quality motions using the quality indicators. Our method can be implemented through a simple diffusion-based framework, leading to a unified motion generate-discriminate model, which can be used to both identify and fix corrupted frames. We demonstrate that our proposed method is effective for training motion cleanup models on raw mocap data in production scenarios by applying StableMotion to SoccerMocap, a 245-hour soccer mocap dataset containing real-world motion artifacts. The trained model effectively corrects a wide range of motion artifacts, reducing motion pops and frozen frames by 68\% and 81\%, respectively. On our benchmark dataset, we further show that cleanup models trained with our method on unpaired corrupted data outperform state-of-the-art methods trained on clean or paired data, while also achieving comparable performance in preserving the content of the original motion clips. 
\end{abstract}

\begin{teaserfigure}
  \vspace{10mm}
  \includegraphics[width=\textwidth]{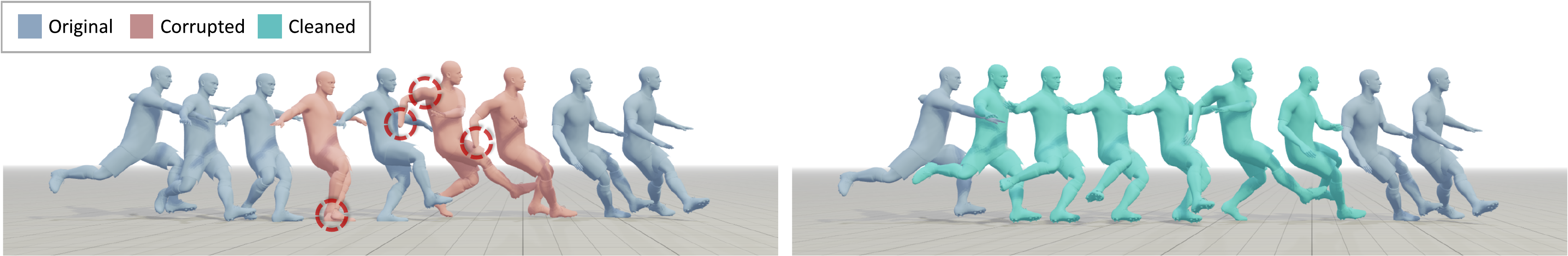}
  \caption{
  StableMotion enables effective training of motion cleanup models using raw mocap data containing real-world artifacts, without requiring paired corrupted-to-clean motion data. Once trained, the model can automatically identify and fix corrupted motions across a wide range of real-world artifacts. This example showcases automatic correction of body self-penetration and unnatural wrist movements. 
  }
  \label{fig:teaser}
  \vspace{5mm}
\end{teaserfigure}


\maketitle

\section{Introduction}

High-quality motion data serves as a cornerstone for numerous applications in computer animation. With the increasing prevalence of deep learning techniques for motion synthesis, the demand for large-scale, high-quality motion datasets has grown significantly~\citep{guo2022generating,peng2022ase,shi2024interactive,tevet2023human,alexanderson2023listen}. However, such datasets are often scarce and challenging to procure. Even motion data captured by professional mocap studios often contain artifacts caused by occlusion, sensor inaccuracies, or errors in post-processing~\citep{holden2018robust}. These issues are further exacerbated when collecting motion data from in-the-wild environments or outdoor scenarios, which are common sources for large-scale datasets~\citep{lin2024motion,jiang2024worldpose}. Cleaning noisy motion data requires significant manual effort from artists, resulting in a costly and labor-intensive process, especially in large-scale production environments that rely on vast amounts of mocap data. 
Data-driven motion cleanup models present a compelling solution by automating the cleanup of noisy data using learned priors from large motion datasets. However, most existing methods for developing motion cleanup models depend on paired datasets of clean and corrupted motions, which are then used to train supervised learning models to correct motion artifacts~\citep{zhang2024rohm,li2025coin,holden2018robust,chen2021mocap}. The creation of such paired training data is inherently challenging, requiring either costly manual cleanup of corrupted data or augmentation with synthetic artifacts on already-cleaned data, which must accurately replicate real-world artifacts. These challenges become especially daunting for large-scale, domain-specific applications, such as sports, where the cost of manual cleanup of these large datasets can quickly become exorbitant.

A similar challenge exists in reinforcement learning (RL), where it is much more costly to collect expert demonstrations compared to suboptimal demonstrations for a specific task. Surprisingly, \citet{kumar2022should} observed that offline RL with suboptimal data could surpass behavior cloning (BC) with expert-only data. 
For instance, Decision Transformer~\citep{chen2021decision}, which models the state-return trajectories, are able to generate novel 
optimal solutions for finding shortest paths in a graph and outperforms BC on Atari games by leveraging diverse suboptimal data.
This performance improvement may stem from the richer diversity inherent in suboptimal data, which enables the policy to learn not only the desired behaviors for a given task, but also which behaviors to avoid~\citep{kumar2022should}. Once the model has been trained, the desired performance of the policy can subsequently be generated by conditioning the policy on the appropriate target return.
Inspired by state-return trajectory modeling in offline RL, we incorporate a frame-level quality indicator variable (\qualvar) similar to state-level reward in RL. Our StableMotion framework utilizes a generate-discriminate approach, where a model is jointly trained to evaluate motion quality and generate motion of varying quality, according to \qualvar. Similar to the practice of prompting text-to-image model with ``photo-realistic quality'', \qualvar offers a knob to specify the generation quality. We further explicitly integrate quality-conditioned motion generation and motion-conditioned \qualvar prediction into a unified diffusion-based model. Once trained, the model can: 1) evaluate the quality of a given motion, and 2) generate high-quality motions by conditioning on the desired quality variables. This allows the model to cleanup raw mocap data by first identifying corrupted segments and then inpainting them with high-quality motions.

The central contribution of this work is StableMotion, a framework for training motion cleanup models from unpaired, corrupted motion data. Instead of learning to directly restore corrupted motions to clean motions, our method leverages all available raw motion data, with mixed quality, using a simple conditional inpainting framework. The trained generate-discriminate model is capable of both evaluating motion quality and generating higher quality motions than those in the original training dataset. Our method can fix corrupted segments by generating higher-quality alternatives, while leaving uncorrupted frames intact. We also propose a number of test-time techniques that harness the sample diversity and the dual functionality of the generate-discriminate model to improve consistency and preservation of the content in the original motion clips. We apply StableMotion to train a cleanup model on a 245-hour, large-scale
soccer mocap dataset, \soccermocap, containing artifacts from real-world capture systems. The model successfully corrects a wide range of motion artifacts, reducing motion pops and frozen frames by 68\% and 81\%, respectively, showcasing StableMotion's efficacy in real-world production scenarios. 

\section{Related Works}
Motion cleanup has long been a crucial challenge in computer animation~\citep{xiao2008automatic}, with impact on a wide range of industries and applications. The recent proliferation of large-scale data-driven methods has amplified the demand for high-quality motion data. However, raw motion capture data inevitably contains artifacts that require cleanup~\citep{holden2018robust,lin2024motion}. Many prior motion cleanup approaches focus on marker-based mocap data, either by directly cleaning raw marker trajectories, or by jointly performing marker-to-skeleton mapping and artifact correction within a unified framework~\citep{holden2018robust,pan2023locality,pan2024romo,chen2021mocap}. However, marker-free systems have become increasingly common for capturing large-scale in-the-wild motion datasets. To support broader synergy with different capture systems, our work targets cleaning of skeleton-based motion data as a more versatile setting~\citep{zhang2024rohm,hwang2025scenemi,li2025coin,mall2017deep}, which is compatible with standard formats, such as BVH, and can be used with data captured from marker-based and marker-free mocap systems.

One approach to cleaning motion artifacts, which often appear as physically implausible behaviors, is to simulate the motion using a physics simulator~\citep{liu2010sampling,zhang2023vid2player3d}. This can improve adherence to physical laws and human-like motor behaviors. However, state-of-the-art physics-based models still often exhibit artifacts, such as jittering and sliding, due to overfitting and inaccurate physics simulations~\citep{luo2023perpetual,yao2024moconvq,tessler2024maskedmimic,peng2018deepmimic,truong2024pdp}. 

Another common approach for motion cleanup involves kinematic-based data-driven methods.
Many kinematic-based regression methods have been proposed to restore clean motions from their corrupted counterparts~\citep{mall2017deep,holden2018robust,pan2024romo,pan2023locality,chen2021mocap,zhang2024rohm,li2025coin}, achieving state-of-the-art performance on motion cleanup tasks. However, these methods require paired corrupted-to-clean data for training, which is often impractical in real production scenarios. Such paired data are often created by synthetically corrupting already-cleaned general-purpose datasets, such as AMASS and LAFAN1~\citep{mahmood2019amass,harvey2020robust}, which requires emulating the characteristics of real-world artifacts. To bypass this challenge, previous methods have also trained models solely on clean datasets and taken advantage of the regularization characteristics of data-driven frameworks to mitigate abnormal behaviors. For instance, autoencoding frameworks can be used to learn compact latent representations of the underlying motion manifolds~\citep{holden2016deep,holden2015learning,aberman2020skeleton}. Once the manifold has been constructed, motion artifacts can be fixed by encoding and decoding through the latent space, leveraging the structure of the latent manifold to regularize and remove artifacts. Similarly, motion synthesis models have also demonstrated promising results for motion cleanup by inpainting corrupted frames with high-quality counterparts using generative models trained on large general-purpose datasets~\citep{cohan2024flexible,shafir2024human,tevet2023human,guo2024momask}. Recent works that combine the advantages of inpainting and the denoising nature of diffusion models have shown more promising results in fixing artifacts while preserving the original content~\citep{hwang2025scenemi}, demonstrating the state-of-the-art performance in post-processing noisy video-based mocap. However, a common limitation of these methods is the need for large-scale, high-quality datasets in order to train models to approximate the manifolds of natural motions~\citep{harvey2018recurrent, harvey2020robust, hernandez2019human, kaufmann2020convolutional}. A model trained on a general-purpose motion dataset may not generalize effectively when applied to domain-specific data, such as sport-related mocap, which are common in production scenarios. On the other hand, acquiring a high-quality in-domain dataset for training domain-specific cleanup models can also be prohibitively expensive, often requiring manual cleanup, and thereby limiting these methods' real-world applicability.

A potentially more practical alternative is to train cleanup models directly from raw corrupted data with mixed quality~\citep{daras2023soft,daras2024ambient,lehtinen2018noise2noise,moran2020noisier2noise,bora2018ambientgan,daras2024much}. However, these techniques are predominantly tailored for the image domain, relying on assumptions about known noise distributions and levels (e.g., Gaussian or salt-and-pepper noise), which are rarely well-modeled in motion data. Our proposed StableMotion framework addresses these challenges by leveraging quality indicator variables (\qualvar) to train motion cleanup model directly from unpaired, corrupted motion data, without requiring prior knowledge of the noise models associated with motion artifacts. A simple interpretation of our method is analogous to a large text-to-image generation model trained on massive internet data, which can be conditioned with quality-related prompts. Similarly, in our framework, quality prompting during motion generation is enabled through quality-aware training using \qualvar on large datasets with mixed quality.

\section{Background: Motion Diffusion Model}
Diffusion models approximate a target data distribution through a progressive diffusion and denoising process. Given a sample $\rvx_0$ from the dataset, 
the forward diffusion process gradually corrupts the sample with Gaussian noise over $T$ steps:
\begin{equation}
    q(\rvx_t|\rvx_{t-1}) = \mathcal{N}(\rvx_t; \sqrt{\alpha_t} \rvx_{t-1}, \beta_t \mathbf{I}),
\end{equation}
where $\beta_t = 1 - \alpha_t$, $t$ denotes the diffusion step, and $\{\alpha_t\}_{t=1}^T$ is a pre-defined noise schedule. This process defines a Markov chain $\{\rvx_t\}_{t=0}^T$, which leads to $q(\rvx_T)$ converging to a Gaussian distribution. 

To approximate the inverse diffusion process for generation, we follow prior motion diffusion models and train a denoising model $g$ to predict the fully denoised sample $\rvx_0$~\citep{tevet2023human}. The denoising model $\hat{\rvx}_0 = g(\rvx_t)$ is trained using the \textit{simple} DDPM objective~\citep{ho2020denoising}:
\begin{equation}
    \mathcal{L}_{\text{simple}} = \mathbb{E}_{\rvx_0 \sim \gM, t \sim \mathcal{U}[1, T], \rvx_t \sim q(\rvx_t|\rvx_0)} \left[ \| \rvx_0 - g(\rvx_t) \|^2 \right].
\end{equation}
\begin{figure}[t]
    \centering
    \includegraphics[width=0.92\linewidth]{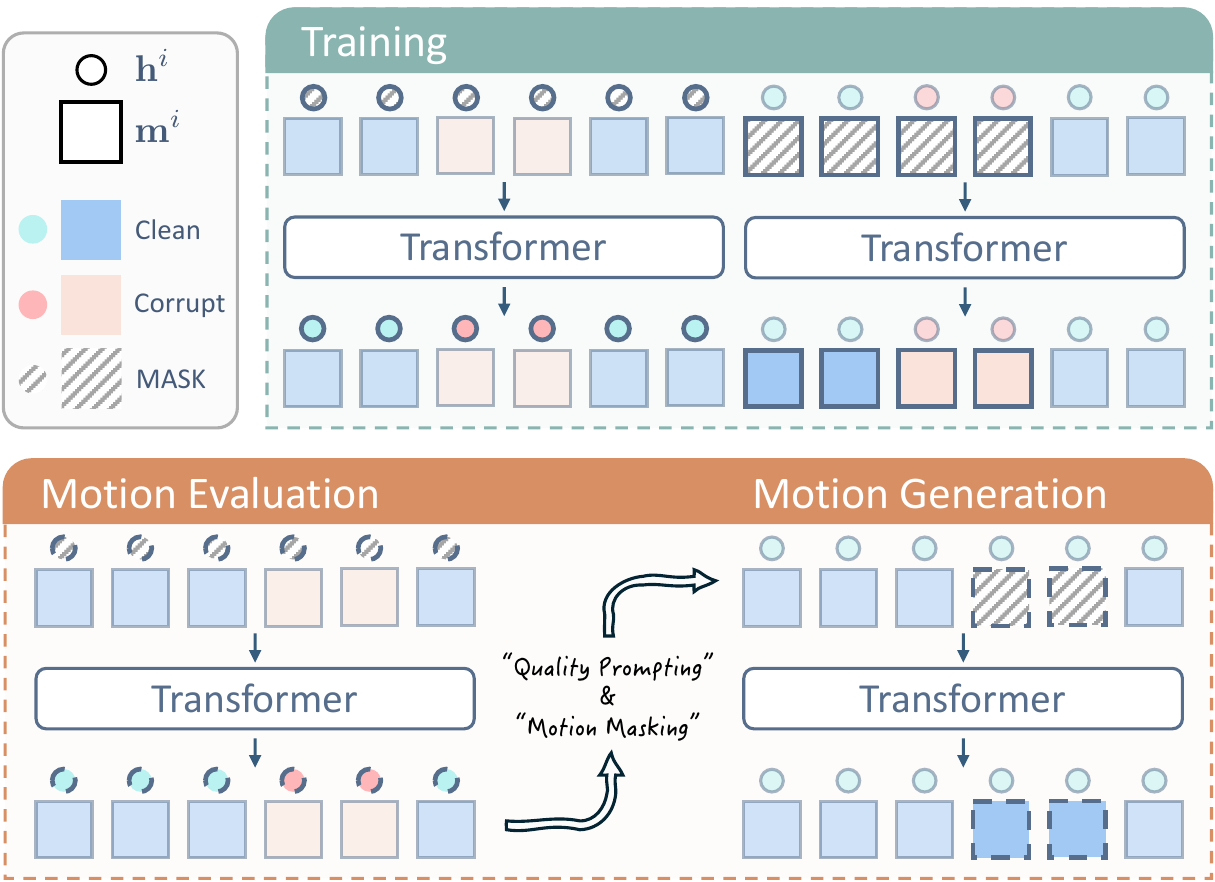}
    \vspace{-2mm}
    \caption{
    \model is built on a diffusion-based generate-discriminate model.
    The model is jointly trained to predict quality indicator variables (\qualvar) $\rvh$ and generate motions $\rvm$. The shaded icons represent features to be inpainted. A key characteristic of our framework is that the model can be directly trained on raw, corrupted motion data of mixed quality. Then at test time, the model can identify corrupted frames and produce high-quality alternatives by ``quality prompting'' through \qualvar.
    }
    \label{fig:pipeline}
\end{figure}

\section{Overview}
\label{sec:overview}
\emph{Problem Definition:} Given a raw skeleton motion sequence that contains artifacts, the objective of a motion cleanup model is to generate a high-quality counterpart that satisfies the following criteria: 1) correct motion artifacts in corrupted frames, and 2) preserve the semantics and content of the original motion.

In this work, our primary goal is to develop a framework for training motion cleanup models directly on raw, domain-specific motion datasets. These models can then be applied to clean artifacts in the desired dataset, resulting in high-quality motion data tailored to the specific domain. Models trained on general-purpose datasets, such as AMASS~\citep{mahmood2019amass}, often fail to generalize effectively to domain-specific data due to differences in depicted behaviors and retargeting errors arising from morphological differences. For instance, the dynamic movements of athletes in sports mocap datasets differ significantly from the everyday motions typically found in general-purpose datasets. In such applications, training cleanup models directly on domain-specific data is vital for effective motion cleanup. However, real-world motion capture data often contain artifacts~\citep{holden2018robust}, which may be randomly distributed throughout motion clips. Constructing paired corrupted-to-clean data for supervised training is both expensive and labor-intensive. Moreover, naively discarding corrupted segments will fragment the dataset into short, disconnected clips (e.g., leading to an average clip length of approximately 1.5s in SoccerMocap), undermining the temporal coherence required for learning long-horizon motion dynamics (see \Cref{tab:soccer} w/ Filtered). These challenges highlight the need for a data-driven framework that can be trained directly on raw mocap data, without breaking the continuity of the underlying sequences.

An overview of StableMotion is illustrated in \Cref{fig:pipeline}. Our generate-discriminate framework unifies the tasks of motion quality evaluation and quality-aware motion generation through a simple inpainting objective (\Cref{sec:model}). During training, the model receives as input a raw motion sequence $\rvm^{1:N}$, along with quality indicators $\rvh^{1:N}$ for each frame, where corrupted frames are marked as $\rvh^i = \mathbf{1}$ and clean frames as $\rvh^i = \mathbf{0}$. The model is trained to jointly generate the motion $\rvm$ of mixed quality and the corresponding \qualvar $\rvh$ through a diffusion-denoising process. 
The trained model can be used to automatically clean motion data by first predicting \qualvar to identify corrupted frames, and then inpainting corrupted frames with higher-quality motions (\Cref{sec:hack}). We propose a number of additional test-time techniques to further improve consistency and preserve the content of the original motion clips.

\section{Generate-Discriminate Model}
\label{sec:model}
The core component of our framework is a generate-discriminate model, which generates motions alongside their corresponding discriminative labels. In this work, the discriminative labels are per-frame quality indicator variables (\qualvar), which enable training quality-aware models from unpaired, corrupted data of mixed quality. The indicator variables also provide a mechanism to direct the model to produce clean motions by conditioning on desired the \qualvar. In this section, we first present the training process of quality-aware diffusion models using unpaired, corrupted motion data. Then, we detail the test-time process of applying the model to identify and clean motion artifacts.

\subsection{Training: Quality-Aware Inpainting}
The standard motion diffusion model provides a strong model for generating motions that follow a given dataset distribution~\citep{tevet2023human}, 
but it cannot selectively generate motions of a desired quality when the dataset contains artifacts. To directly train a motion cleanup model on uncleaned raw mocap data, we introduce quality indicator variables $\rvh$, to represent the underlying distribution of different data quality levels. 
We train our model directly on the combined features of the motion and its corresponding \qualvar, $\rvx_{t} = (\rvm, \rvh)_{t}$, where $t$ denotes the diffusion step. 
By incorporating \qualvar into the feature representation, the motion diffusion model can represent motions of varying quality with explicit control over the quality of the generated motions. This simple approach enables effective training of motion diffusion models on raw corrupted motion data, which can then generate high-quality clean motion.

To further tailor the model for motion cleanup, we design two inpainting training tasks 
to facilitate the key functions of our motion cleanup model: 1) motion quality evaluation and 2) quality-aware motion generation.
At runtime, the model is provided with either motion frames or \qualvar, and the model then infers the missing features. 
While prior methods have shown that an unconditional motion diffusion model $g(\rvx_t)$ can perform similar inpainting tasks using inference-time imputation techniques~\citep{lugmayr2022repaint,tevet2023human}, recent work has demonstrated that explicitly training a conditional inpainting model tends to outperform methods relying solely on test-time imputation strategies with an unconditional model~\citep{cohan2024flexible,karunratanakul2023guided}. 
Based on these observations, we adopt a conditional model for motion cleanup. Given an inpainting mask $\rvb$, where observed conditional features are marked as 1 and inpainting features as 0,  a quality-aware motion diffusion model can be trained using a masked inpainting objective. The mask is randomly generated in two modes: 1) motion evaluation and 2) motion generation. For motion evaluation, the goal is to infer the quality indicator variables $\rvh^{1:N}$ for all frames, given the motion sequence $\rvm^{1:N}$. For motion generation, we denote the missing frame indices as $i \in \sA \subseteq \{1, \dots, N\}$ and the observed frame indices as $i \in \sB = \sA^\complement$. The goal is to generate the missing motion frames $\rvm^{i \in \sA}$ given the observed frames $\rvm^{i \in \sB}$ and the per-frame specified quality indicators $\rvh^{1:N}$.
Both modes can be easily implemented through different inpainting masks, providing a unified quality-aware inpainting objective.

\subsection{Inference: Motion Evaluation and Generation}
\label{sec:infer}
As illustrated in \Cref{fig:pipeline}, by specifying different test-time inpainting masks $\rvb$, the model trained with our proposed generate-discriminate approach can function as either a motion quality evaluator $D$ or a motion generator $G$. 

\textit{Motion Quality Evaluation}:
For motion quality evaluation, the inpainting mask obscures the \qualvar for each frame, and the model is then required to infer the corresponding \qualvar for each frame $D(\rvh|\rvm)$. 
The predicted \qualvar can then be used to detect corrupted segments or evaluate the overall quality of a motion sequence.

\textit{High-Quality Motion Generation}:
For motion generation, the model approximates the distribution of missing frames conditioned on observed frames and quality indicators, serving as a motion inpainting model $G(\rvm^{i \in \sA} | \rvm^{i \in \sB}, \rvh^{1:N})$.
During inference, the \qualvar can all be set to $\rvh \gets \mathbf{0}$, which prompts the model to generate high-quality motions for all missing frames.

\section{Motion Cleanup}
\label{sec:hack}
Once the generate-discriminate model has been trained, it can be used to clean raw mocap data by identifying corrupted frames and inpainting them with clean motions, while leaving the uncorrupted frames unchanged. As described in \Cref{sec:infer}, the model can function as a motion quality evaluator $D(\rvh|\rvm)$. Given a raw mocap sequence $\rvm$ as input, the model evaluates each frame to determine if a frame exhibits artifacts. Frames that require cleaning are identified as $\sA \gets \{ i \mid D(\rvh^{i \in 1:N}|\rvm) = \mathbf{1} \}$. The input for motion inpainting is then constructed by masking out the segments $(\rvm, \rvh)^{i \in \sA}$ and assigning \qualvar as $\rvh \gets \mathbf{0}$, which specifies high-quality motions for all frames. The model is then used to generate clean motion for all corrupted frames $G(\rvm^{i \in \sA} | \rvm^{i \in \sB}, \mathbf{0}^{1:N})$.
In our experiments, we observe that the generate-discriminate model exhibits strong motion cleanup capabilities, but can lead to drastic changes in the content of the original motion in some scenarios due to the masked-inpainting process. Furthermore, the inherent stochasticity of diffusion models can occasionally produces suboptimal results. In this section, we describe a number of test-time techniques designed to address these limitation and further improve the consistency of our system.

\begin{figure}[t]
    \centering
    \includegraphics[width=\linewidth]{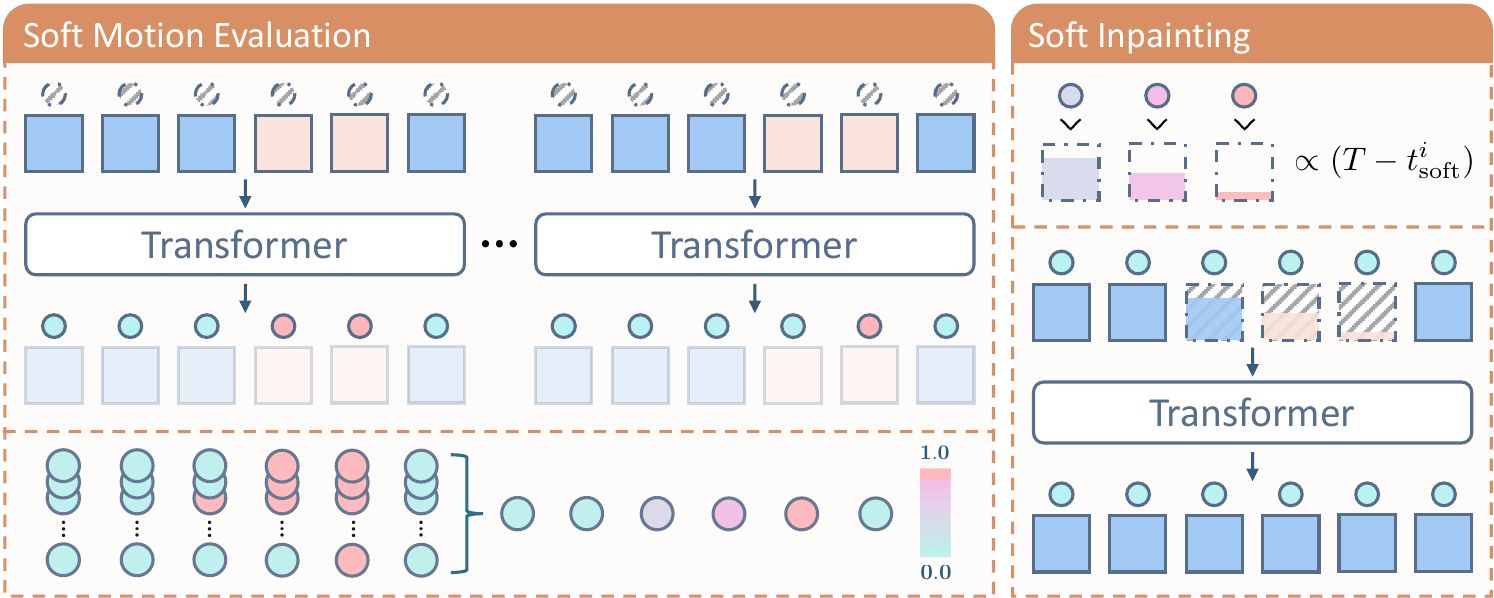}
    \vspace{-5mm}
    \caption{
    To improve preservation of motion content, we propose an adaptive cleanup technique using soft motion quality evaluation and soft motion inpainting. A simple Monte Carlo method is used to estimate \emph{soft} quality labels $\bar{\rvh}$ from the predicted \qualvar $\rvh$, which approximates the severity of artifacts in each frame. Then, we design a soft inpainting strategy that inpaints corrupted frames by initializing the denoising process for each frame $\rvm^i$ at different diffusion steps ${\rvt^{i}_\text{soft}}$. When correcting frames with subtle artifacts (i.e. low $\bar{\rvh}^i$), the inpainting process is initialized at later denoising steps $\rvt^{i}_\text{soft} \textless T$, thereby retaining more information from the original frames.
    }
    \label{fig:softi_pipeline}
\end{figure}

\begin{figure}[t]
\centering
\includegraphics[width=0.85\linewidth]{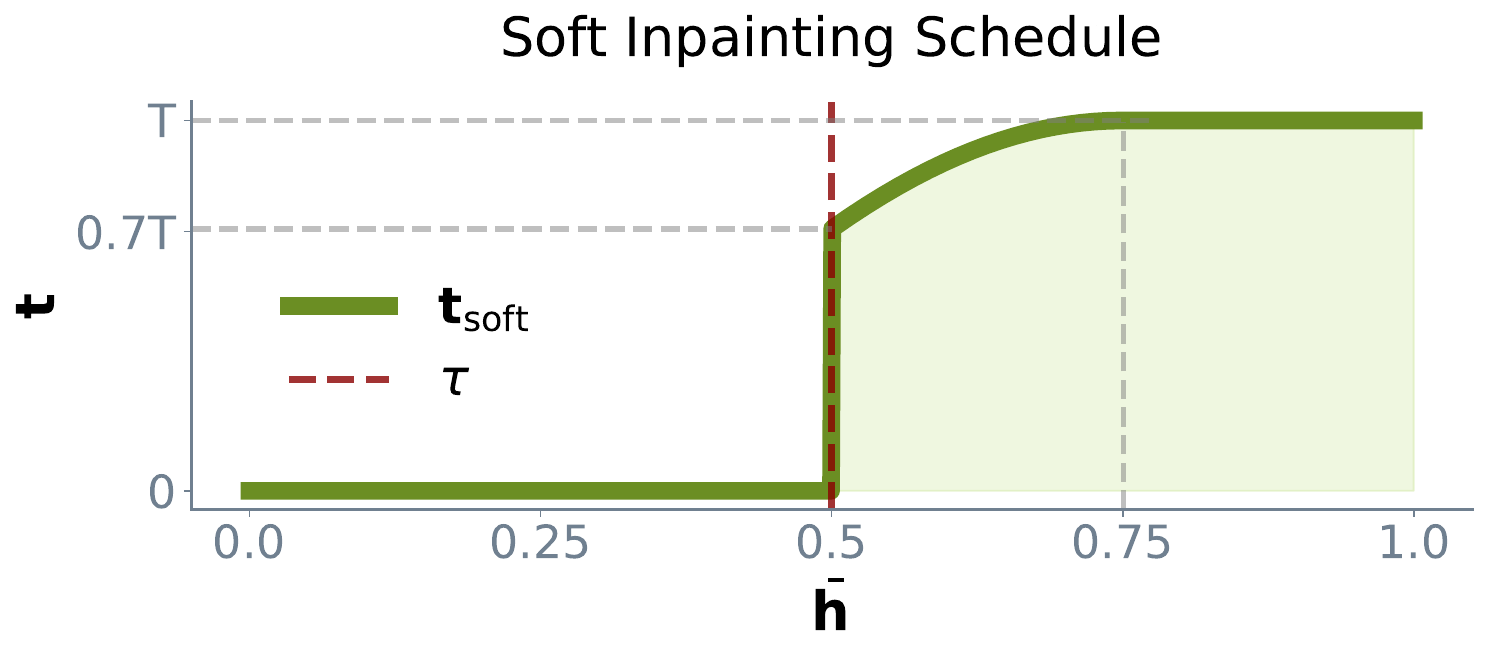}
\caption{The soft inpainting schedule initializes inpainting from different diffusion steps based on the estimated soft quality label of each frame, $\bar{\rvh^{i}}$. Frames with lower $\bar{\rvh}$, indicating fewer artifacts, are inpainted from later diffusion steps to retain more original information. Frames with higher $\bar{\rvh}$, indicating severe artifacts, are inpainted from earlier steps to better remove the corruption while preserving overall motion coherence.}
\label{fig:softi_schedule}
\end{figure}

\subsection{Adaptive Cleanup}
\label{subsec:adaclean}
During inference, the standard diffusion sampler initializes $\rvx_T$ with Gaussian noise, which introduces stochasticity to the inpainting process. This random initialization also erases all features of the original corrupted frames, even if the original frames contained only subtle artifacts and still retain useful information. To adaptively erase and inpaint corrupted frames based on their level of corruption, we propose an adaptive cleanup technique, shown in \Cref{fig:softi_pipeline}, which takes advantage of the denoising process of a diffusion model and the dual functionality of our generate-discriminate model.

To identify the level of corruption in each frame, we approximate a \emph{soft} quality label $\bar{\rvh}^i$ by taking multiple samples of $\hat{\rvh}^i$ for each frame, and then computing the expected value over the samples. This approach provides a \emph{soft} estimate of the level of corruption in each frame, instead of relying directly on the value of a single sample. This approximation assumes that severely corrupted frames are more consistently identified as being corrupted, resulting in higher expected values of the indicator variable, while frames with more subtle artifacts tend to yield lower expected values.

Once the level of corruption for each frame has been estimated, we use the diffusion-denoising process to generate motions starting from different intermediate diffusion steps $\rvt^i_{\text{soft}}$ at each frame rather than the maximum step $T$.
The initial denoising step $\rvt^i_{\text{soft}}$ is determined based on the soft quality labels $\bar{\rvh}^i$. In practice, we observe that starting from later denoising steps (\textit{e.g.}, $\rvt_{\text{soft}} < \frac{T}{2}$) often fails to fully remove artifacts.
To address this, we design a soft inpainting schedule that encourages inpainting from earlier diffusion steps, while still retaining useful information from the original corrupted frames. The initial denoising step is determined according to:
\begin{equation}
\small
    \rvt^i_{\text{soft}} = 
    \begin{cases}
        T\sin{\frac{\pi}{2}\min \left(1,2 \bar{\rvh}^i-1+\tau \right)}, & \text{if } \bar{\rvh}^i \geq \tau \\
        0, & \mathrm{otherwise} \\
    \end{cases} ,
\end{equation}
where $\bar{\rvh}^i \approx \mathbb{E}[\hat{\rvh}^i]$ denotes the Monte-Carlo estimate of the soft quality label for the $i$-th frame, and $\tau = 0.5$ is the threshold to determine whether a frame should be inpainted. A visualization of the soft inpainting schedule is illustrated in \Cref{fig:softi_schedule}. With this schedule, soft inpainting is performed when $\bar{\rvh}^i$ lies between $0.5$ and $0.75$, while for $\bar{\rvh}^i > 0.75$, the full denoising process will be applied starting from diffusion step $T$. This adaptive cleanup technique allows the model to adaptively adjust the modification to each frame based on the severity of the artifacts, enabling it to better preserve the content of the original frames.

\begin{figure}[t]
    \centering
    \includegraphics[width=\linewidth]{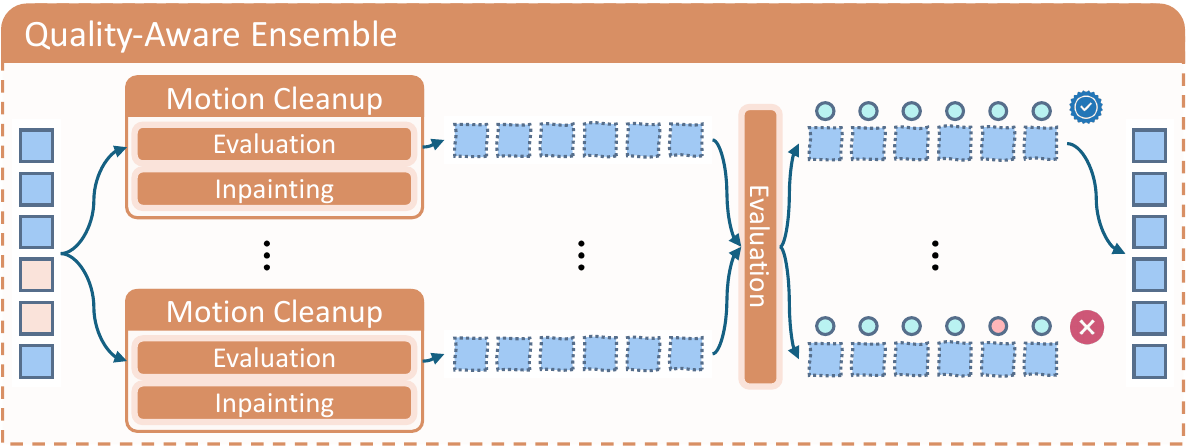}
    \caption{
    The quality-aware ensemble leverages the diversity of diffusion models and the dual functionality of generate-discriminate models to boost motion cleanup performance. This approach ensembles diverse candidate motions by selecting the highest-quality motion based on predicted motion quality scores, leading to more consistent and higher-quality results than performing a single pass of the cleanup model.
    }
    \label{fig:ensemble_pipeline}
    \vspace{-5mm}
\end{figure}

\begin{figure*}[t]
    \centering
    \includegraphics[width=\linewidth]{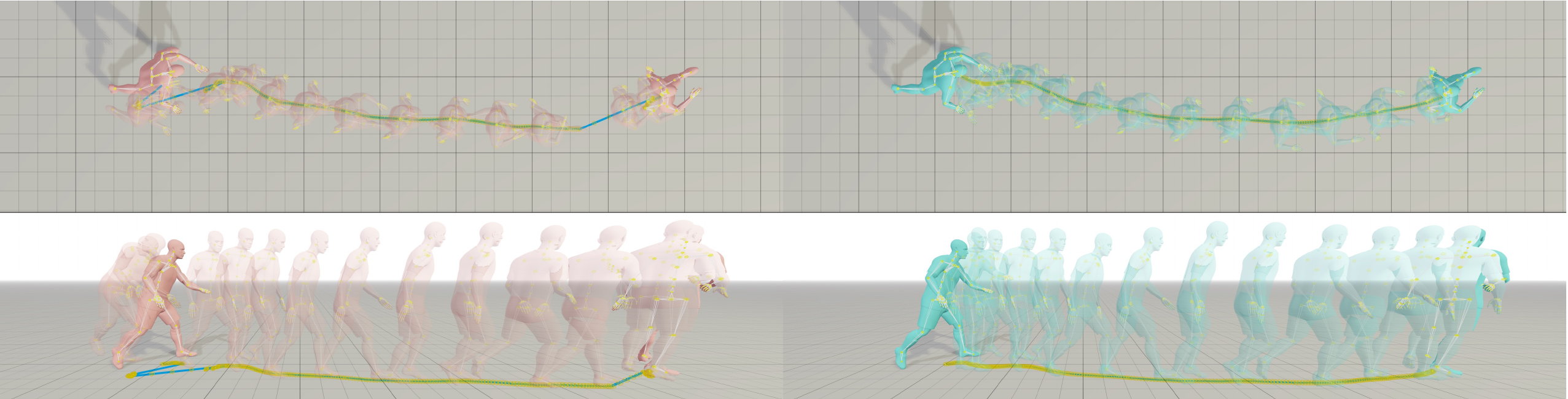}
    \caption{
    Example of fixing frozen frames and motion pops in SoccerMocap. These artifacts can take hours for a professional animator to fix, while StableMotion can correct them smoothly in a few seconds.
    }
    \label{fig:soccerpop}
\end{figure*}

\subsection{Quality-Aware Ensemble}
To further improve the consistency and performance of our system, we leverage the sample diversity of diffusion models through ensembling, where results from multiple diverse models are combined to improve generalization~\citep{kotu2019data}. However, instead of generating samples from different models, we propose to ensemble different predictions from a single model. An overview of this ensemble process is illustrated in \Cref{fig:ensemble_pipeline}. We first execute the motion cleanup process multiple times, consisting of motion evaluation and motion inpainting, and collect the candidate output motions. Since each DDPM inference process starts with independent Gaussian noise, the outcomes are diverse and independent. However, naively averaging the results is unsuitable for motion data, as averaging across diverse modes can introduce artifacts and distort the natural dynamics of the motion. Instead, the generate-discriminate model enables selection among the candidates based on the estimated motion quality. The motion evaluation procedure $D(\rvh|\rvm)$ is applied to each candidate motion to estimate the quality of each candidate. The overall quality score of a motion sequence is computed as the sum of $\rvh$ at each frame and provides an automatic criterion for selecting the best quality candidate. In our experiments, we generate 5 candidate motions for ensembling. Since multiple motions can be generated in parallel using batch inference, this approach does not significantly increase the overall wall-clock time of the motion cleanup process. More detailed ablation experiments that evaluate the effectiveness of our quality-aware ensembling approach is available in \Cref{subsec:inference_ablation}.

\section{Model Representations}
In this section, we describe the implementation details for several crucial components of the system.

\paragraph{Motion Representation.}
The commonly used relative trajectory representation can lead to error accumulation when reconstructing global motions~\citep{holden2016deep,tevet2023human}. To localize edits and preserve uncorrupted frames during cleanup, we adopt a global motion representation inspired by prior work on human mesh recovery and motion cleanup~\citep{zhang2024rohm,hwang2025scenemi}. With this global representation, each frame is represented by features consisting of: global root translation, ground-projected facing direction, local joint rotations and positions, global foot positions and velocities, and global root velocity. Here, ``global'' features are defined relative to the first frame, which is canonicalized to ensure consistent initial orientation. This results in a 244-dimensional feature vector $\rvm^i$ per frame for SoccerMocap, which is further augmented with a 1D quality indicator $\rvh^i$.

\paragraph{Network Architecture.} 
Our model is implemented using a diffusion transformer (DiT)~\citep{peebles2023scalable}. Adaptive normalization is used to condition the model on the diffusion step and a class label~\citep{chen2024pixartalpha},
which indicates whether the motion or \qualvar is observed during inpainting. To better capture relative temporal correlations between frames, we adopt rotary positional embeddings (RoPE) in the attention layers~\citep{su2024roformer,shen2024world}. More architectural details are available in \Cref{sec:appendix_arch}.

\begin{figure*}[t]
    \centering
    \includegraphics[width=\linewidth]{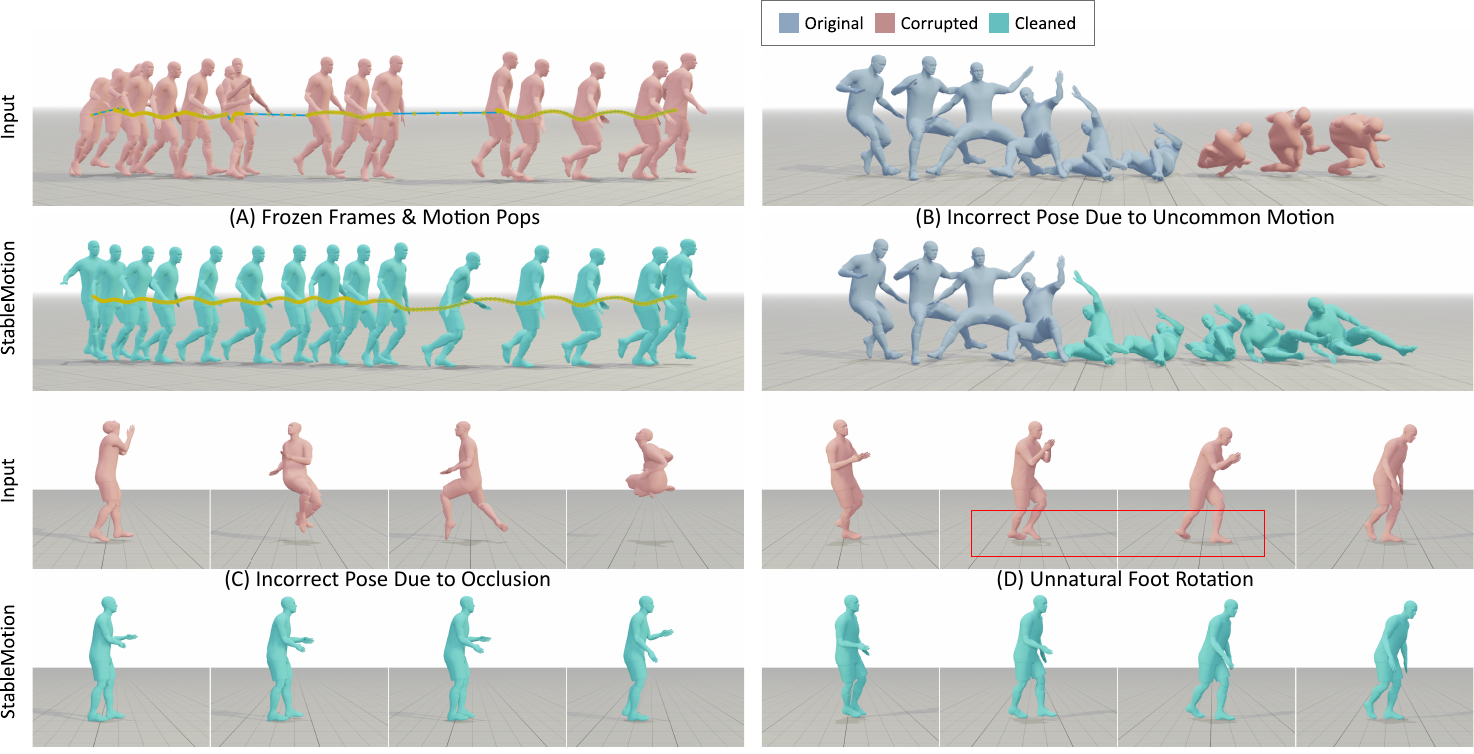}
    \caption{
    Examples of corrupted motion clips in SoccerMocap before and after clean up by our model. \textbf{(A)} Fixing segments of frozen frames and motion pops while preserving the root trajectory. \textbf{(B)} Fixing corrupted motions on the ground. \textbf{(C)} Fixing abrupt rotations and incorrect joint positions caused by camera occlusion. \textbf{(D)} Fixing subtle unnatural foot rotations.
    }
    \label{fig:soccerdemo}
\end{figure*}

\section{Real-world Experiments: SoccerMocap}
\label{sec:soccermocapexp}
We start by evaluating the effectiveness of our framework on SoccerMocap, a real-world mocap dataset from a prominent game studio. 
\emph{SoccerMocap} is a proprietary, large-scale soccer dataset captured using optical and marker-free motion capture systems in full-field environments. It contains approximately 245 hours of motion data recorded at 50 fps, and exhibits artifacts introduced by in-the-wild capture conditions, such as large camera-to-subject distances, occlusions, system failures, and equipment vibrations. 
Since the majority of corrupted segments are much shorter than 5 seconds, the maximum context length of our model is set to 5 seconds. Once trained, long motion clips can be cleaned by applying our model with a sliding window.

The quality indicator variables used for training are annotated by a joint effort of artists and heuristic algorithms. Frames containing artifacts are first annotated by artists, accounting for approximately 1\% of the frames. These annotations focus on challenging artifacts, such as body self-penetration and abnormal pose errors, which can be difficult to detect using heuristic algorithms. We refer to these \textbf{h}uman-\textbf{a}nnotated \textbf{a}rtifacts as \textbf{HAA} in the following discussion. To expand the annotations, we apply heuristic artifact detection algorithms designed to identify foot-skating, flipping, motion pops, and frozen frames. This process results in approximately 39\% of the frames being identified as corrupted. These heuristics are designed to achieve high recall at the expense of lower precision compared to human annotations, ensuring that nearly all corrupted frames are correctly identified. The dataset is split into a 240-hour training set and a 5-hour testing set.

\begin{figure*}[t]
    \centering
    \includegraphics[width=\linewidth]{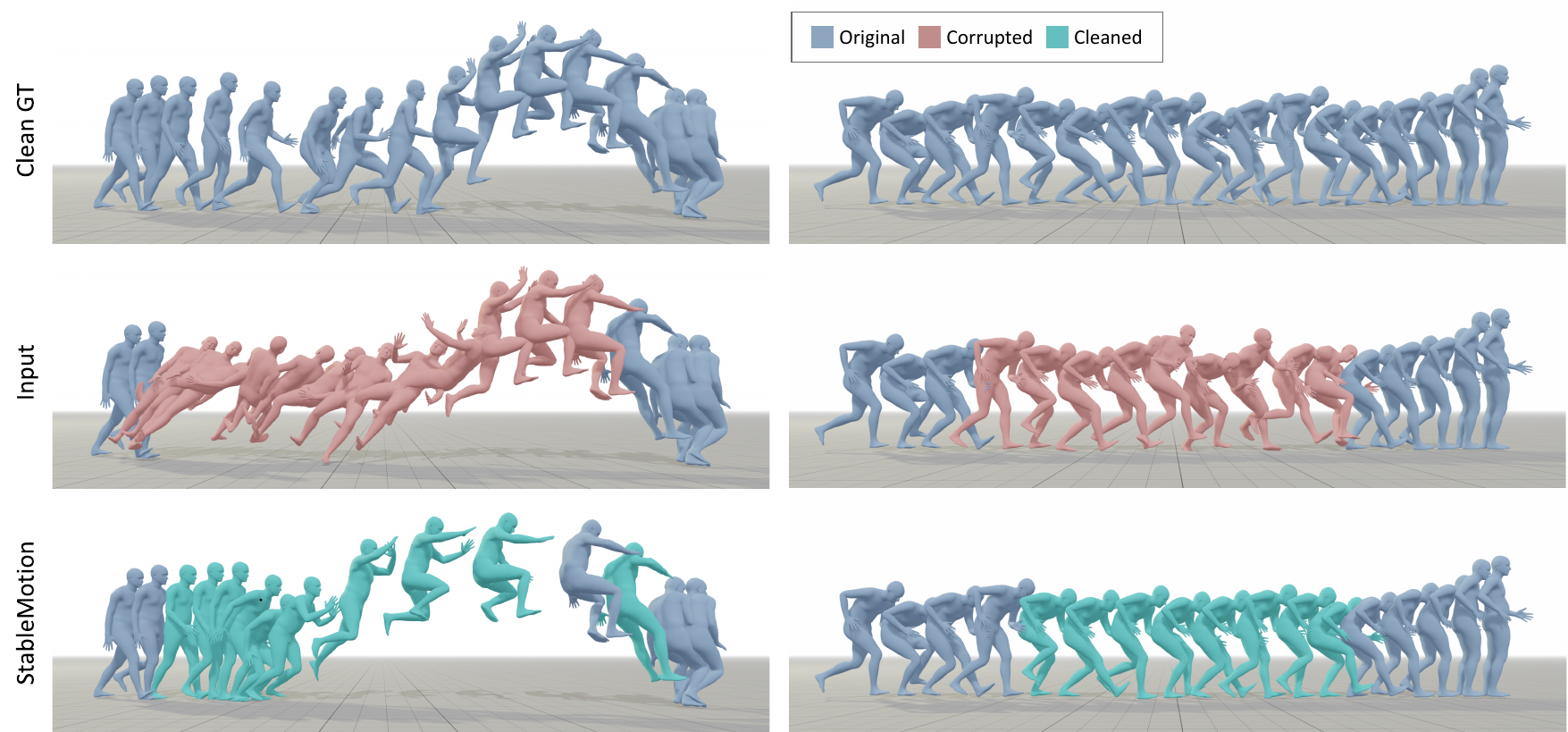}
    \caption{
    Locomotion clips in BrokenAMASS before and after cleanup by our model. Our method demonstrates strong performance in cleaning up dynamic locomotion behaviors, producing natural foot and body movements while preserving the global trajectory.
    }
    \label{fig:amassdemo1}
\end{figure*}

\paragraph{Results} 
StableMotion models are trained directly on the SoccerMocap training set, and then applied to clean the testing set. Examples of motions from SoccerMocap before and after cleanup by the model are illustrated in \Cref{fig:teaser} and \Cref{fig:soccerdemo}. For the kicking motion shown in \Cref{fig:teaser}, it can take 30--60 minutes for a professional animator to fix the body self-penetrations, whereas our model detects and corrects it automatically with reasonably natural and consistent results. Our model is able to automatically remove artifacts such as frozen frames and motion pops, shown in \Cref{fig:soccerdemo} (A) and \Cref{fig:soccerpop}, which can take hours for animators to fix. The model has been used to automatically process newly captured soccer mocap data from similar systems. 
The kicking motion shown in \Cref{fig:teaser} is an example from a recent capture session.

\begin{table}[t]
    \caption{Performance of the StableMotion on SoccerMocap. A proprietary artifact detection algorithm is used to evaluate motion quality before and after cleanup. \textbf{FS Dist} measures the severity of foot skating. \textbf{Pops Rate} quantifies the frequency of jittery transitions and motion pops. \textbf{Frozen Rate} captures the frequency of over-smoothed transitions and frozen frames. 
    StableMotion provides a highly effective method for training motion cleanup models directly on raw, corrupted datasets in real-world scenarios. 
    }
    \label{tab:soccer}    
    \centering
    \resizebox{\linewidth}{!}{
    \begin{tabular}{l|c|c|c|c}
        \toprule
        \textbf{Method} & \textbf{FS Dist} $\downarrow$ & \textbf{Pops Rate} $\downarrow$ & \textbf{Frozen Rate} $\downarrow$ & \textbf{Jittering} $\downarrow$ \\
        \midrule
        Input & 2.39 & 0.41 \% & 23.86 \% & 8.94 \\
        \midrule
        XClean & 2.86 &  0.35 \% &  23.07 \% & 15.18 \\
        ConvAE-B & 9.54 &  47.46 \% &  5.07 \% & 104.44 \\
        CondMDI-B & 3.25 &  18.21 \% &  4.64\% & 22.09 \\
        \midrule
        w/o \qualvar & 3.76 & 1.91 \% & 6.43 \% & 9.85 \\
        w/ Filtered & 2.96 & 0.23 \% & \cellcolor{green!10}\textbf{4.43} \% & 14.67 \\
        \model & \cellcolor{green!10}\textbf{2.14} & \cellcolor{green!10}\textbf{0.13} \% & 4.50 \% & \cellcolor{green!10}\textbf{8.68} \\
        \bottomrule
    \end{tabular}
    }
    \vspace{-2mm}
\end{table}

Since previous methods require either clean or paired corrupted-to-clean data to train cleanup models, most cannot be directly trained on \soccermocap{}, a raw mocap dataset with pervasive artifacts. 
To evaluate StableMotion's effective utilization of raw motion data, we compare StableMotion against two ablations of our method: one trained only on filtered clean data (w/ Filtered), and the other trained on mixed-quality data without using \qualvar{} (w/o \qualvar). To compare StableMotion with a traditional non-learning based cleanup technique, we implement an example-based cleanup method (XClean)~\citep{Examplehumanmotion,motiongenerationfromexamples}, commonly used in game studio pipelines. We also compare our method with a state-of-the-art motion inpainting model, CondMDI~\citep{cohan2024flexible}, trained on raw, mixed-quality data without text conditions, which we refer to as CondMDI-B. To compare with inpainting-based methods developed prior to the rise of diffusion models~\citep{harvey2018recurrent, harvey2020robust, hernandez2019human, kaufmann2020convolutional}, we constructed a convolutional autoencoder baseline following \citet{kaufmann2020convolutional}, denoted as ConvAE-B, and reimplement it using a more modern network architecture.
All baseline models use the same motion representation. Since not all baseline methods have the ability to identify corrupted frames, the corrupted frame labels predicted by our model is used for all baseline methods. To quantitatively evaluate performance, we use a proprietary motion artifact detection algorithm to estimate artifact severity before and after cleanup. As shown in \Cref{tab:soccer}, \textbf{FS Dist}, \textbf{Pops Rate}, \textbf{Frozen Rate}, and \textbf{Jittering} measure the levels of foot skating, motion pops, frozen frames, and jittery behaviors, respectively. 
We refer the reader to \Cref{sec:appendix_metrics} for more details about these metrics.

\begin{figure*}[t]
    \centering
    \includegraphics[width=\linewidth]{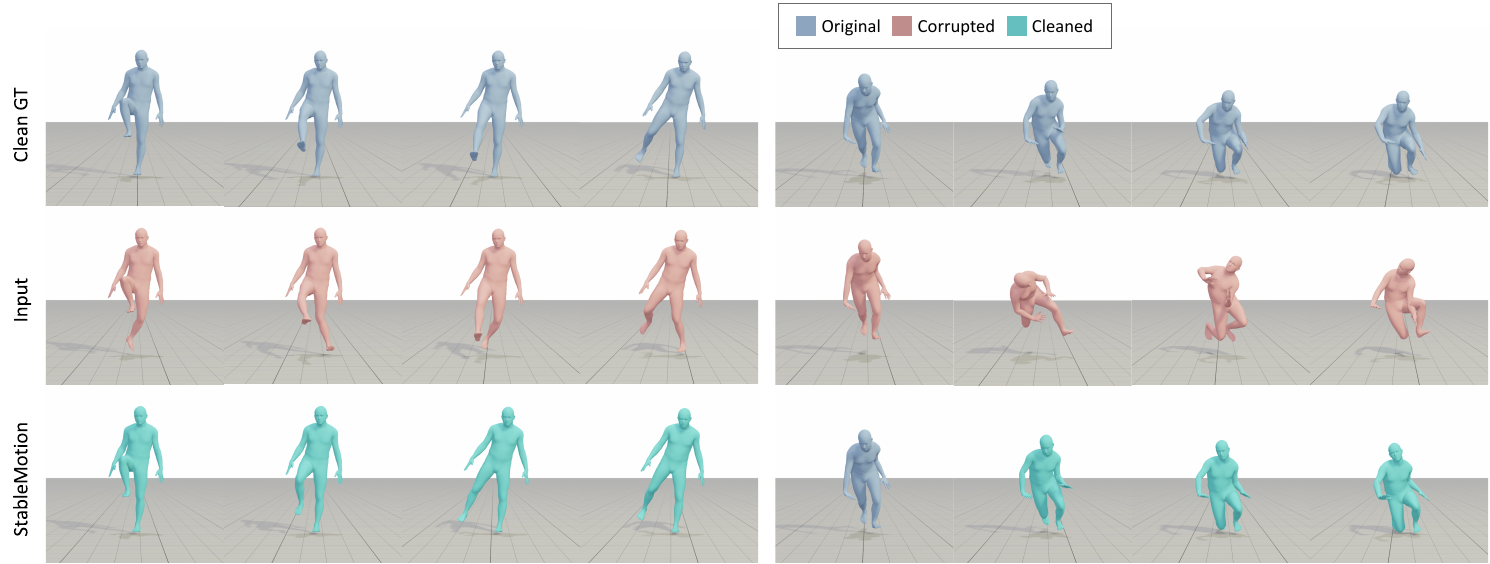}
    \caption{
    Examples of corrupted motion clips in the BrokenAMASS dataset before and after cleanup by our model. The model built with StableMotion effectively fixes motion artifacts, such as jittering and unnatural poses.
    }
    \label{fig:amassdemo2}
\end{figure*}

ConvAE, CondMDI-B, and the ``w/o \qualvar'' baseline show that directly training models on raw data can exhibit, or even amplify, motion artifacts—potentially due to the influence of artifacts, such as jittery or popping, in the training data. The results from ``w/ Filtered'' further show that simply filtering out corrupted frames and training the motion generation component of our model on the resulting clean but fragmented snippets can degrade motion quality, possibly due to the loss of temporal coherency. In contrast, StableMotion effectively trains a motion cleanup model directly on raw data using the quality indicators, preserving the continuity and structure of the motion sequences. The model is able to correct a wide range of motion artifacts introduced by real-world mocap systems and exhibits more effective performance than the baselines. To evaluate our model's accuracy in identifying artifacts, we compare the predicted quality indicators with the quality labels from human annotators. Our model accurately detect subtle artifacts that typically require manual annotation, achieving an \textbf{HAA} recall rate of 81.54\%.

\section{Benchmark Experiments: BrokenAMASS}
To evaluate the effectiveness of StableMotion for training cleanup models from corrupted data in a controlled setting, we create BrokenAMASS by introducing artifacts into the AMASS dataset to simulate real-world raw mocap data that contains artifacts without clean paired motions~\citep{mahmood2019amass}. This dataset contains over 40 hours of motion data from the original AMASS dataset, downsampled to recorded at 20 fps. The dataset covers diverse motion types, including locomotion, transitions, sports, dance, yoga, and martial arts~\citep{BABEL:CVPR:2021}.
We design a collection of corruption functions, including random-joint perturbation, over-acceleration, drifting, and over-smoothing, to simulate real-world motion artifacts, such as jittering, foot sliding, and other unnatural movements. Detailed implementations of these corruption functions are provided in the Appendix. The default BrokenAMASS dataset contains approximately 23\% corrupted frames. The dataset is split into training, test, and validation sets with a 0.80 : 0.15 : 0.05 ratio.

\paragraph{Metrics.}
BrokenAMASS serves as a controlled benchmark for evaluating content preservation by enabling direct comparison between the cleaned-up motions produced by different models and the corresponding clean ground-truth motions from AMASS~\citep{mahmood2019amass}. Following prior work, we also adopt several heuristic metrics that estimate the severity of common motion artifacts, such as foot skating and jittering. The evaluation metrics include: \textbf{FS Dist} for foot sliding distance~\citep{ling2020character}, \textbf{Jitter} from the first derivative of acceleration~\citep{shen2024world}, and \textbf{Accel} for acceleration error compared to clean motion~\citep{zhang2024rohm}. \textbf{GMPJPE} assesses joint position errors in global frames. \textbf{M2M Score} and \textbf{M2M R@3} quantify motion content semantic similarity using the TMR model~\citep{petrovich2023tmr}. \textbf{GMPJPE} and \textbf{M2M} provide complementary metrics to evaluate content preservation and high-level semantic consistency respectively.

\begin{table}[t]
\caption{
Quantitative comparison on BrokenAMASS. Results are averaged over three random seeds and reported with standard deviations. The best performing model is highlighted, and \underline{underlines} denote the second-best. ``$\rightarrow$'' means that values closer to the ground truth are better. Methods marked with ``*'' use ground-truth quality labels to identify frames for cleanup. Our method effectively removes motion artifacts while also preserving the original motion content. 
}
\vspace{-2mm}
\label{tab:comparison}
\centering
\resizebox{\linewidth}{!}{
\begin{tabular}{l|c|c|c|c|c|c}
\toprule
\textbf{Method} & \small \textbf{FS Dist} $\downarrow$ & \small\textbf{Jitter} $\rightarrow$ & \small\textbf{Accel} $\downarrow$& \small\textbf{GMPJPE} $\downarrow$ & \small\textbf{M2M Score} $\uparrow$ & \small\textbf{M2M R@3} $\uparrow$\\ 
\midrule
Clean GT & 3.85  & 1.26 & 0.00 & 0.000  & 1.000 & 100.0 \\ 
Input & 6.42 & 7.32 & 2.03 & 1.39  & 0.946 & 88.8 \\ 
\midrule
ConvAE-B* & \et{5.08}{0.08} & \et{6.12}{0.33} & \et{2.18}{0.07} & \et{9.07}{0.17}   & \et{0.883}{0.005}  & \et{66.0}{1.36} \\ 
CondMDI-B* & \et{3.96}{0.02} & \et{2.32}{0.03} & \et{1.25}{0.01} & \et{9.17}{0.01}   & \et{0.879}{0.000}  & \et{63.2}{0.22} \\ 
CondMDI* & \ets{3.84}{0.01} & \et{2.26}{0.04} & \et{1.22}{0.01} & \et{9.14}{0.02}   & \et{0.883}{0.000} & \et{63.9}{0.32} \\ 
RoHM & \et{4.43}{0.11} & \ets{2.02}{0.15} & \ets{1.00}{0.03} & \ets{7.56}{0.63}   & \cellcolor{green!10}\etb{0.940}{0.002} & \ets{81.2}{0.54} \\ 
StableMotion  & \cellcolor{green!10}\etb{3.70}{0.01} & \cellcolor{green!10}\etb{1.14}{0.04} & \cellcolor{green!10}\etb{0.60}{0.01} & \cellcolor{green!10}\etb{2.42}{0.04}  & \ets{0.938}{0.003} & \cellcolor{green!10}\etb{84.5}{0.77} \\ 
\bottomrule
\end{tabular}
}
\vspace{-2mm}
\end{table}

\begin{figure}[t]
\centering
\includegraphics[width=\linewidth]{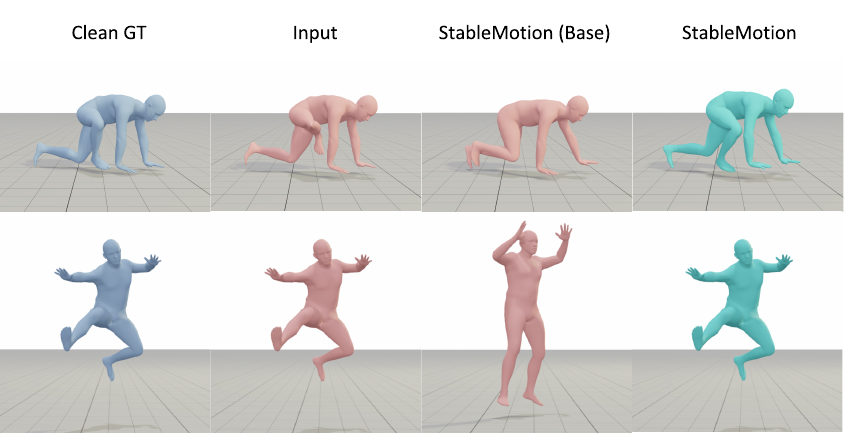}
\caption{
StableMotion model with the Adaptive Cleanup (\Cref{subsec:adaclean}) produces higher-quality and more consistent results compared to the vanilla inference setting (Base), particularly in preserving motion content when artifacts are subtle. Our inference design enhances consistency and achieves improved results, especially for rarely seen actions that cannot be easily inferred from neighboring frames. Additional ablation experiments for our inference techniques are available in the supplementary material.
}
\label{fig:vsbase}
\end{figure}

\paragraph{Comparisons.}
We compare the performance of our method against strong baseline methods from prior work, each trained under their intended dataset conditions using either \textit{clean} or \textit{paired corrupted-to-clean} motion data: (1) CondMDI, a conditional motion in-betweening diffusion model trained on clean data~\citep{cohan2024flexible}; and (2) RoHM, a non-inpainting diffusion-based motion restoration method trained with paired corrupted-to-clean data~\citep{zhang2024rohm}. To ensure a fair comparison, we train each baseline on the type of data that best matches its original design. 
Since existing methods are not designed to be trained directly on raw, corrupted motion data, we construct both a diffusion-based and a non-diffusion-based motion in-betweening model trained on mixed-quality unpaired data, which we denote as CondMDI-B and ConvAE-B~\citep{kaufmann2020convolutional}.
As shown in \Cref{tab:comparison}, CondMDI trained on corrupted data performs worse than the version trained purely on clean data. This performance gap indicates that the CondMDI framework relies on training with clean data to be effective, limiting its applicability in scenarios where obtaining clean datasets is challenging. In contrast, our model is trained only on unpaired corrupted data, but still outperforms state-of-the-art cleanup models that are trained with cleaned motion data. The motions corrected by our model exhibit even fewer foot skating artifacts compared to the clean motions from the original AMASS dataset. Moreover, our method is able to better preserve the content of the original motions by applying the cleanup process only to the corrupted frames. This is particularly crucial for highly dynamic locomotion behaviors, as illustrated in \Cref{fig:amassdemo1}. Compared to RoHM, the non-inpainting model trained on paired data, our method demonstrates stronger performance for fixing artifacts, achieving a 16\% reduction in foot skating distance and a 40\% reduction in acceleration error. StableMotion also achieves comparable performance in content preservation, with similar semantic preservation performance and a 68\% reduction in global joint position errors.
    
\paragraph{Qualitative Results.}
As shown in \Cref{fig:amassdemo2}, our model effectively fixes motion artifacts such as jittering and unnatural poses across a diverse range of motions. Furthermore, as illustrated in \Cref{fig:amassdemo1}, when the initial frames of a jumping motion are highly corrupted, our model is able to restore the corrupted frames with a natural transition between the running and jumping phases of the motion. Our test-time techniques, including adaptive cleanup and quality-aware ensembling, can further improve cleanup performance, particularly in preserving the original motion content, as illustrated in \Cref{fig:vsbase}. Qualitative comparisons with baseline methods are available in the supplementary video.

\begin{figure}[t]
    \centering
    \includegraphics[width=0.9\linewidth]{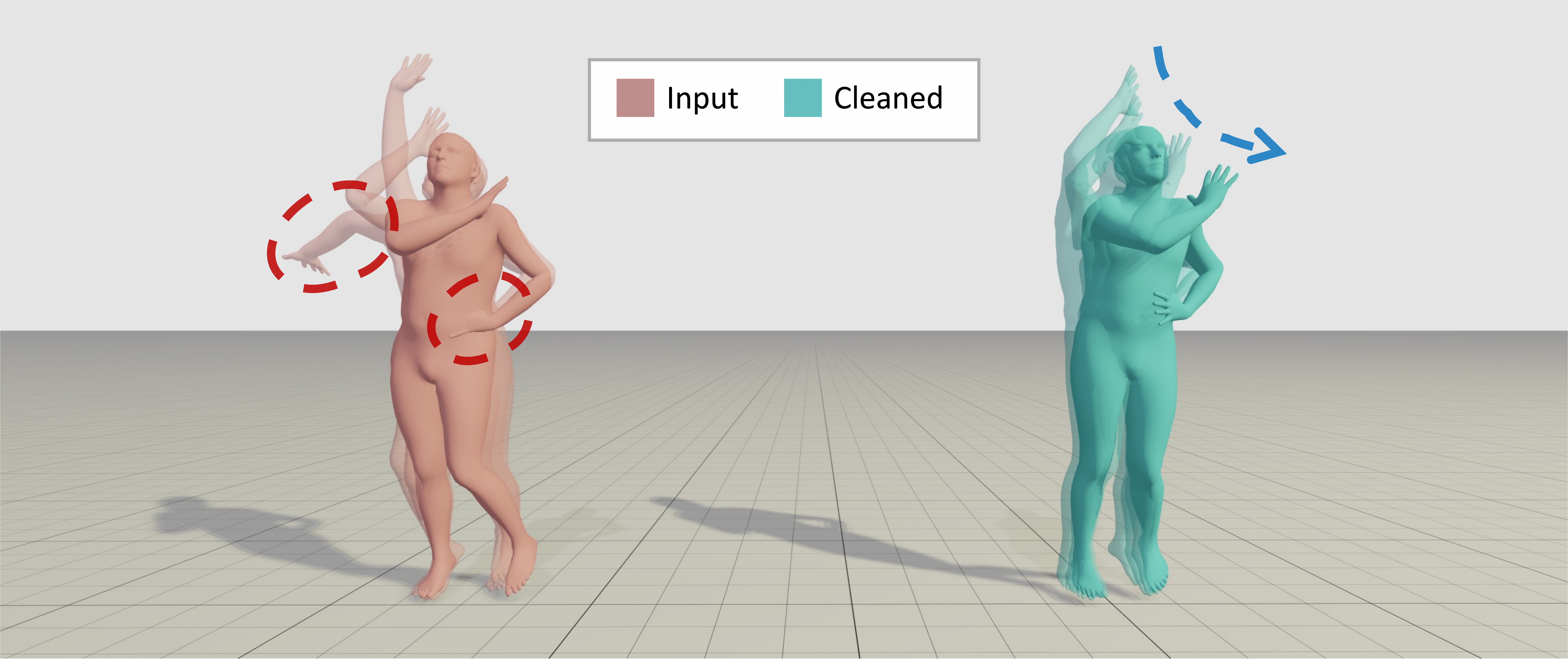}
    \caption{
    StableMotion is able to fix highly jittery motions from PopDanceSet.
    }
    \label{fig:popdance}
\end{figure}

\section{Real-world Experiments: PopDanceSet}
To further demonstrate the application of our domain-specific motion cleanup method to in-the-wild datasets, we apply StableMotion to PopDanceSet~\citep{luo2024popdg}, a dance motion dataset collected from web videos using pose estimation algorithms. 
It contains over 3 hours of diverse dance motions from more than 100 subjects, recorded at 30 fps and represented in the SMPL format~\citep{loper2015smpl}. 
However, due to the inherent challenges of 3D pose estimation from monocular video—video-based mocap datasets often exhibits low motion quality and contains numerous jittery artifacts, as reported in \Cref{tab:popdance}. We adopt similar evaluation metrics as the BrokenAMASS experiments. However, since clean ground-truth motions are unavailable, metrics such as \textbf{Accel} and \textbf{GMPJPE} are computed with respect to the raw input motions. To further assess the severity of different artifacts in real-world mocap data, we include additional metrics, such as \textbf{Foot Skating Rate (FS Rate)} and \textbf{Foot Penetration Distance (FP Dist)}.

We compare our method against both a diffusion-based and a non-diffusion-based approach, denoted as CondMDI-B and ConvAE-B, following the implementations from the previous experiments. All models are trained on the PopDanceSet training split and evaluated on the test split. The quality labels for training StableMotion are annotated using automatic heuristics.
During testing, since not all baselines are able to detect corrupted frame, we use the quality labels predicted by StableMotion for all methods to ensure a fair comparison. As shown in \Cref{tab:popdance}, despite being trained on a highly corrupted dataset, our model is able to effectively artifacts while preserving the original motion content. An example of cleanup motion jittering using StableMotion is shown in \Cref{fig:popdance}.
The non-diffusion-based method ConvAE-B is also able to effectively reduce jitter artifacts, which may be due to deterministic models being more robust to jittery artifacts that resemble Gaussian noise~\citep{lehtinen2018noise2noise}. However, these deterministic models tend to struggle for more challenging dataset, such as the highly dynamic SoccerMocap dataset or BrokenAMASS, where the prevailing artifacts may not resemble Gaussian noise.

\begin{table}[t]
\caption{
Quantitative comparison on PopDanceSet. Results are averaged over three random seeds and reported with standard deviations. The best performing model is highlighted.}
\label{tab:popdance}
\centering
\resizebox{\linewidth}{!}{
\begin{tabular}{l|c|c|c|c|c|c}
\toprule
\textbf{Method} & \small \textbf{FS Dist} $\downarrow$ & \small \textbf{FS Rate} $\downarrow$ & \small \textbf{FP Dist} $\downarrow$ &  \small\textbf{Jitter} $\downarrow$ & \small\textbf{Accel} $\downarrow$& \small\textbf{GMPJPE} $\downarrow$\\ 
\midrule
Input & 6.41  & 17.62 \% & 12.35 & 27.97 & 0.00  & 0.00 \\ 
\midrule
ConvAE-B & \et{4.78}{0.11} & \et{12.47}{1.98} \% & \et{8.30}{1.04} & \et{15.87}{0.13} & \et{8.51}{0.01} & \et{3.71}{0.05}\\ 
CondMDI-B & \et{5.87}{0.05} & \et{15.28}{0.42} \% & \et{8.56}{0.17} & \et{22.59}{0.15} & \et{9.27}{0.03} & \et{3.95}{0.02}\\ 
StableMotion  & \cellcolor{green!10}\etb{4.52}{0.04} & \cellcolor{green!10}\etb{10.03}{0.38} \% & \cellcolor{green!10}\etb{6.82}{0.13} &  \cellcolor{green!10}\etb{15.05}{0.05} & \cellcolor{green!10}\etb{7.76}{0.01} & \cellcolor{green!10}\etb{3.13}{0.04}\\ 
\bottomrule
\end{tabular}
}
\end{table}

\section{Discussion and Future Work}
In this work, we present StableMotion, a novel framework for training motion cleanup models directly on unpaired, corrupted motion datasets. StableMotion employs a diffusion-based generate-discriminate model that simultaneously evaluates motion quality and generates motions of varying quality through the use quality indicator variables. Our model can be trained with a simple masked inpainting strategy, and can then effectively clean a wide range of motion artifacts without relying on training with clean motion data, which is often costly and time-consuming to acquire. Experiments on SoccerMocap demonstrate StableMotion's effectiveness in large-scale real-world production datasets. 

Our framework demonstrates that a clean motion dataset is not necessarily a prerequisite for developing effective motion cleanup models, opening potential opportunities of training models on large-scale, raw motion data commonly found in animation and related fields. However, the inpainting procedure utilized in our framework can lead to drastic changes of the content of a motion clip in some scenarios. For applications with higher motion precision requirements, such as dexterous manipulation, test-time classifier guidance may be incorporated to better preserve the content of the original motions. Another notable limitation of our framework is its reliance on motion quality labels for training, which can still involve a considerable amount of manual effort to annotate. However, we show in \Cref{sec:appendix_dataablation} that effective motion cleanup models can be trained when only a small portion of the dataset is annotated with quality labels. Developing automated techniques that can accurately identify motion artifacts can further improve the effectiveness of our framework and expand its application across more diverse domains.

\begin{acks}
This work was supported by NSERC (RGPIN-2015-04843) and the National Research Council Canada (AI4D-166). We would like to thank the anonymous reviewers for their constructive comments, Alessandro Sestini, Chuan Guo, and Shihao Zou for their support for this work.
\end{acks}

\bibliographystyle{ACM-Reference-Format}
\bibliography{main}

\clearpage
\appendix

\section*{Table of Contents}
\startcontents
\printcontents{}{1}{}

\section{Evaluation Metrics Details}
\label{sec:appendix_metrics}

In this section, we describe all evaluation metrics involved in our experiments. The evaluation metrics include:
\begin{itemize} \small
    \item \textbf{Foot Skating Distance (FS Dist)} and \textbf{Foot Skating Rate (FS Rate)}: Measure the extent and frequency of foot sliding on the floor. The computation of FS Dist follows the implementation in \citet{ling2020character}.
    \item \textbf{Foot Penetration Distance (FP Dist)}: Measures the extent of foot-ground penetration.
    \item \textbf{Jittering Measurement (Jitter)}: Approximates the jitteriness of a motion using the first derivative of acceleration~\citep{shen2024world}.
    \item \textbf{Acceleration Error (Accel)}: Measures the difference in acceleration between cleaned-up motions and ground-truth 3D joints~\citep{zhang2024rohm}.
    \item \textbf{Frozen Rate} and \textbf{Pops Rate}: ``Frozen frame'' artifacts refer to segments where a character or specific joints remain unnaturally static for several consecutive frames, breaking the expected fluidity of motion. These often result from tracking failures, occlusions, or post-processing errors. ``Motion pops'' describe sudden, unnatural discontinuities or abrupt jumps in joint positions or orientations between frames, producing visually jarring transitions. We use a proprietary artifact detection algorithm, developed in a game studio production workflow, to compute \textbf{Frozen Rate} and \textbf{Pops Rate}, which quantify the frequency of frozen frames \& over-smoothed transitions~\citep{rempe2021humor}, and motion pops, respectively.
    \item \textbf{Global Mean Per Joint Position Error (GMPJPE)}: Evaluate joint position errors in the global coordinate system, with respect to the clean ground-truth motion when available, or the input motion otherwise.
    \item \textbf{Motion-to-Motion Semantic Similarity Score (M2M Score)} and \textbf{Recall at rank 3 (M2M R@3)}: Computed using a pretrained text-motion retrieval model from TMR~\citep{petrovich2023tmr}; R@3 computes the percentage of times the correct motion is among the top 3 results.
\end{itemize}
The first set of metrics, such as \textbf{FS Dist}, \textbf{FP Dist}, and \textbf{Jitter}, aim to quantify the overall quality of the cleaned-up motion~\citep{ling2020character, shen2024world}. These metrics focus on evaluating issues like foot skating, ground penetration, and jittery movements. \textbf{Accel} evaluates the quality of the cleaned-up motion with respect to the original clean motion clips~\citep{zhang2024rohm}. The latter metrics, including \textbf{GMPJPE} and \textbf{M2M Score}, measure how well the model preserves the content of the original motion clips, as the clean-up model should modify the content of the original motion as little as possible when removing artifacts.

We further provide implementation details of several important metrics. 

\textbf{Foot Skating Distance.}
We calculate the Foot Skating Distance (FS Dist) following the implementation in \citet{ling2020character}, defined as:
\begin{equation}\small
    \textbf{FS Dist} := \frac{1}{N} \sum_{n=1}^{N} \left( \lVert \rvf_{n+1} - \rvf_n \rVert \cdot \left( 2 - 2^{\frac{\bar{\rvf}_n}{H}} \right) \cdot \mathds{1}\left(\bar{\rvf}_n < 0.05 \land \rvy_{n} > 0.65\right) \right),
\end{equation}
where $\rvy$ is the height of the root, $\rvf$ is the 2D projection of the foot position on the ground, $\bar{f}_t = \frac{f_t + f_{t+1}}{2}$ is the average foot height at two consecutive frames.

\textbf{Foot Skating Rate.}
We compute the Foot Skating Rate (FS Rate) following the approach described in \citet{zhang2024rohm}:
\begin{itemize}
\item \textit{Skating Conditions}: Skating is determined based on predefined thresholds for velocity and height. Following \citet{zhang2024rohm}, the velocity threshold is set to 0.10 m/s, while the height thresholds are 0.10 m for the toe and 0.15 m for the ankle.
\item \textit{Skating Ratio}: The FS Rate is calculated as the average proportion of timesteps where both the left and right foot meet the \textit{skating conditions}.
\end{itemize}

\textbf{Motion-to-Motion Semantic Similarity.} 
We use the pretrained motion semantic retrieval model TMR to encode motion into latent space and calculate the semantic similarity in the latent space, following \citet{petrovich2023tmr}. Since TMR is trained on AMASS primarily for text-to-motion retrieval, its latent space encodes text-aligned motion semantic information and generalizes well on BrokenAMASS, which shares a similar motion source. This enables it to effectively evaluate semantic similarity between the clean ground-truth motion and the cleanup motion, as similarly applied in \citet{athanasiou2024motionfix}. The M2M Score is computed in the latent space with cosine similarity. Furthermore, retrieval recall at rank 3 (M2M R@3) is evaluated across the entire test set. These two metrics evaluate how well the model preserves the content of the original motion clips.

\begin{figure}[t]
    \centering
    \includegraphics[width=\linewidth]{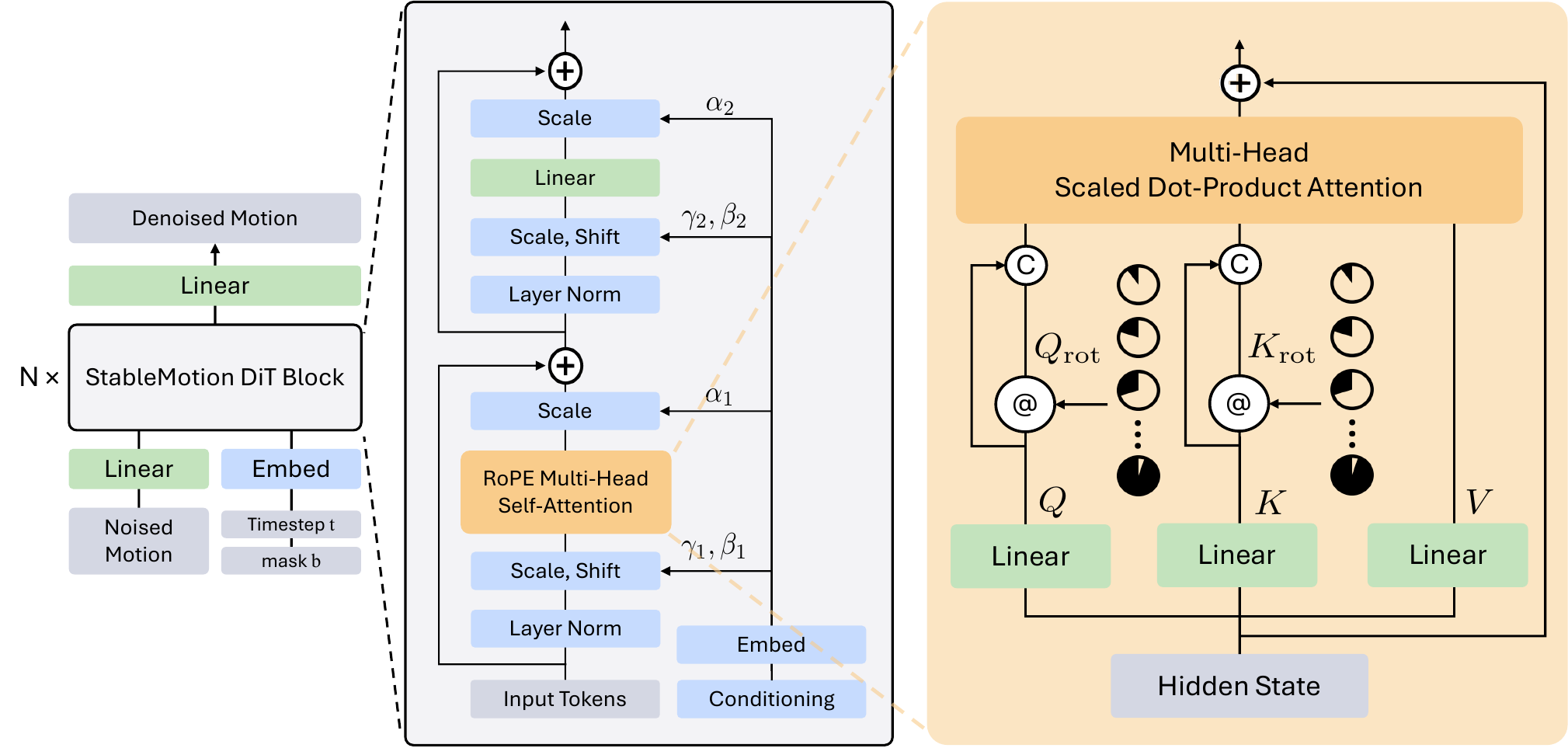}
    \caption{
    Our StableMotion model is implemented using a diffusion transformer~\citep{peebles2023scalable}. The key modifications we introduce is the incorporation of rotary positional embeddings (RoPE) instead of absolute positional embeddings, which are applied to the hidden vectors $Q$ and $K$ in the multi-head self-attention module. The operation ``$@$'' denotes application of a rotation to the vectors. The final key/query used for scaled dot-product attention is constructed by concatenating the original $Q/K$ with their rotated counterparts, $Q_\text{rot}/K_\text{rot}$.
    }
    \label{fig:arch}
\end{figure}

\textbf{Recall at Human-Annotated Artifacts (R@HAA).}
As described in \Cref{sec:soccermocapexp}, challenging artifacts in SoccerMocap, such as body penetration and subtle pose errors that are difficult for automatic algorithms to detect, are manually annotated by artists. We calculate our model's artifact detection recall of these human-annotated artifacts (HAA). R@HAA is calculated following the recall definition:
\begin{equation}
    \textbf{R@HAA} := \frac{TP_{\text{HAA}}}{TP_{\text{HAA}} + FN_{\text{HAA}}},
\end{equation}
where $TP_{\text{HAA}}$ counts the correctly predicted corrupted frames that are both annotated by human annotators and detected by our model, and $FN_{\text{HAA}}$ counts the corrupted frames annotated by human annotators but not detected by our model. This metric measures the sensitivity of the model to hard-to-detect artifacts.

\section{Network Architecture} 
\label{sec:appendix_arch}
Our StableMotion model is implemented using a diffusion transformer (DiT)~\citep{peebles2023scalable}. A schematic diagram of the network architecture used for the denoising model is illustrated in \Cref{fig:arch}. 
Adaptive normalization is used to condition the model on the diffusion step $t$~\citep{perez2018film}, a technique that has shown strong performance in previous diffusion model architectures~\citep{peebles2023scalable, chen2024pixartalpha, dhariwal2021diffusion}. 
Adaptive normalization is also used to incorporate a class label derived from the inpainting mask $\rvb$, which indicates whether the motion $\rvm$ or quality indicator variables $\rvh$ are observed during inpainting. 
This class label is used to switch between motion generation and evaluation modes, allowing the model to seamlessly adapt its functionality to the given input.
To better capture relative temporal correlations between motion frames, we adopt rotary positional embeddings (RoPE) in the attention layers~\citep{su2024roformer,shen2024world,evans2024stable},
instead of absolute positional embeddings.

\section{Motion Representation Details} 
\label{sec:representationdetails}
A commonly used motion representation in previous work is to record the full-body pose with respect to the local root coordinate frame and to record the root trajectory, including translation and rotation, relative to the previous frame~\citep{holden2016deep, guo2022generating}. However, this relative root representation can lead to significant error accumulation when recovering global motion, since the global root trajectory must be reconstructed by integrating the displacement at each frame. 
Furthermore, modifications to corrupted frames can lead to changes in the overall global trajectory of subsequent frames. Our framework aims to localize changes only to corrupted frames while preserving clean frames as much as possible. Therefore, we adopt a global motion representation similar to \citet{cohan2024flexible}, where each frame of the motion is represented using the following features:

\begin{itemize}
    \item Global translation of the root,
    \item Ground-projected facing direction vector,
    \item Local rotation of each joint,
    \item Local position of each joint,
    \item Global position of each foot,
    \item Global velocity of each foot,
    \item Global velocity of the root,
\end{itemize}
where ``Global'' features are defined relative to the first frame, which is canonicalized to ensure a consistent initial facing direction across all sequences. 

The global root translation and local joint rotation are the primary motion features that can be readily extracted from raw BVH skeleton motion files. This also allows the cleaned-up motion to be easily written back into common industry-standard formats, such as BVH.

\section{Training Data Characteristics}
\label{sec:appendix_dataablation}
The crucial component of StableMotion that enables cleanup models to be effectively trained with the unpaired corrupted datasets is the introduction of motion quality indicator variables. In the following experiments, we investigate two key aspects of the corrupted motion dataset: 1) the portion of the dataset that is annotated with quality indicator variables; 2) the ratio of corrupted frames in the dataset. To control the training data characteristics, experiments are conducted on BrokenAMASS.

\textbf{Scaling Quality Labels.} 
Our proposed framework provides a method for training a motion cleanup model without the need to construct a clean motion dataset. Instead, the model learns directly from raw mocap datasets containing artifacts, using binary quality indicator variables (\qualvar) to discern between clean and corrupted frames. Annotating these binary quality labels can require substantially less manual effort compared to fixing artifacts in the motion data. However, annotating these binary quality labels still requires some effort, whether manual or automated. 
To evaluate the impact of the quantity of quality labels, we train the StableMotion models on datasets of varying sizes that are annotated with quality labels and evaluate the models on the BrokenAMASS test set. As shown in \Cref{tab:data_scale}, the model trained on just 10\% of the original dataset already demonstrates strong performance. 
Furthermore, we observe some scaling effects: as the dataset size increases, the model achieves progressively better performance across most of the metrics. When trained on the full dataset, the StableMotion model exhibits the best performance in preserving the semantic content of motion, potentially due to the richer semantic priors learned from the larger dataset. These results suggest that the StableMotion model can be successfully trained when only a small portion of the dataset is annotated with quality labels, enabling it to effectively clean the entire dataset.

\begin{table}[t]
    \caption{Performance of StableMotion model when trained with different sized subsets of BrokenAMASS with quality labels. All results are reported from StableMotion (Base) inference setting without the proposed special test-time techniques.}
    \label{tab:data_scale}    
    \centering
    \resizebox{\linewidth}{!}{
    \begin{tabular}{l|c|c|c|c|c}
        \toprule
        \textbf{Dataset Size} & \textbf{FS Dist} $\downarrow$ & \textbf{FS Rate (\%)} $\downarrow$ & \textbf{Jitter} $\rightarrow$ & \textbf{Accel} $\downarrow$ & \textbf{M2M Score} $\uparrow$ \\
        \midrule
        Clean GT & 3.85 & 3.45 & 1.26 & 0.00 & 1.000 \\
        Input & 6.42 & 13.0 & 7.32 & 2.03 & 0.946 \\
        \midrule
        10\%  & \et{3.59}{0.04} & \et{4.99}{0.28} & \et{1.39}{0.07} & \et{0.70}{0.01} & \et{0.908}{0.002} \\
        25\%  & \et{3.53}{0.04} & \et{3.62}{0.08} & \etb{1.23}{0.07} & \et{0.68}{0.02} & \et{0.902}{0.001} \\
        50\%  & \et{3.46}{0.03} & \et{3.19}{0.14} & \et{1.15}{0.00} & \et{0.66}{0.00} & \et{0.904}{0.005} \\
        100\%  & \etb{3.45}{0.01} & \etb{2.74}{0.03} & \et{1.10}{0.03} & \etb{0.59}{0.01} & \etb{0.926}{0.003} \\
        \bottomrule
    \end{tabular}
    }
    
\end{table}

\textbf{Ablation Corruption Rate.} 
In the BrokenAMASS dataset, approximately 23\% of the frames are corrupted. To investigate the impact of the corruption ratio of the training data on StableMotion, we stress test the model using a highly corrupted dataset, where over 50\% of the motion frames contain artifacts. To isolate the effect of predicting the quality indicator variable and focus on the model's ability to generate clean motions from corrupted data, the statistics reported in \Cref{tab:corrupt_rate} are computed using ground truth quality labels. Even when more than half of the training data are corrupted, StableMotion demonstrates strong performance in fixing artifacts and preserving the content of the original motion clips.

\begin{table}[t]
    \caption{Effect of training data corruption rate on StableMotion framework. ``*'' marks that results are obtained using the ground truth quality labels to specify the segments for cleanup. This setting focuses on the model's high-quality motion generation ability learned from the corrupted data, excluding the effect of the quality indicator variable prediction pefromance.}
    \label{tab:corrupt_rate}
    \centering
    \resizebox{\linewidth}{!}{
    \begin{tabular}{l|c|c|c|c|c}
        \toprule
        \textbf{Corrupted Rate} & \textbf{FS Dist} $\downarrow$ & \textbf{FS Rate (\%)} $\downarrow$ & \textbf{Jitter} $\rightarrow$ & \textbf{Accel} $\downarrow$ & \textbf{M2M Score} $\uparrow$ \\
        \midrule
        Clean GT & 3.85 & 3.45 & 1.26 & 0.00 & 1.000 \\
        Input & 6.42 & 13.0 & 7.32 & 2.03 & 0.946 \\
        \midrule
        Corrupted-23\%* & \etb{3.60}{0.01} & \etb{3.60}{0.12} & \et{1.10}{0.01} & \etb{0.58}{0.00} & \et{0.927}{0.002} \\
        Corrupted-40\%* & \et{3.68}{0.01} & \et{4.04}{0.11} & \et{1.10}{0.02} & \et{0.59}{0.00} & \et{0.928}{0.001} \\
        Corrupted-57\%* & \et{3.72}{0.01} & \et{4.23}{0.11} & \etb{1.13}{0.01} & \et{0.60}{0.00} & \etb{0.927}{0.000} \\
        \bottomrule
    \end{tabular}
    }
\end{table}

\section{Ablations}
\label{sec:appendix_ablation}

\subsection{Ablations on Model Design}
We conduct ablation experiments to evaluate the impact of two key design components: quality indicator variables and rotary positional embeddings (RoPE). 

\begin{table}[t]
    \caption{Ablation experiments on quality indicator variables (\qualvar) and rotary positional embeddings (RoPE). Methods marked with “*” use the ground truth quality labels to determine which frames need cleanup.}
    \label{tab:appendix_ablation}
    \centering
    \resizebox{\linewidth}{!}{
    \begin{tabular}{l|c|c|c|c|c}
        \toprule
        \textbf{Method} & \textbf{FS Dist} $\downarrow$ & \textbf{FS Rate (\%)} $\downarrow$ &\textbf{FP Dist} $\downarrow$ & \textbf{FP Rate (\%)} $\downarrow$  & \textbf{Accel} $\downarrow$ \\
        \midrule
        Clean GT & 3.85 & 3.45 & 4.10 & 1.51 & 0.00 \\
        Input & 6.42 & 13.04 & 4.77 & 1.76 & 2.03  \\
        \midrule
        StableMotion (Base) & \etb{3.45}{0.01} & \etb{2.74}{0.03} & \et{3.73}{0.11} & \etb{1.28}{0.03} & \etb{0.59}{0.01} \\
        w/o RoPE & \et{3.54}{0.03} & \et{3.24}{0.19} & \etb{3.73}{0.07} & \et{1.30}{0.02} & \et{0.62}{0.01} \\
        \midrule
        StableMotion (Base)* & \etb{3.60}{0.01} & \etb{3.60}{0.12} & \etb{3.67}{0.06} & \etb{1.17}{0.02} & \etb{0.58}{0.00} \\
        w/o RoPE* & \et{3.71}{0.02} & \et{4.31}{0.28} & \et{3.70}{0.06} & \et{1.19}{0.01} & \et{0.62}{0.01} \\
        w/o \qualvar* & \et{3.94}{0.02} & \et{6.37}{0.01}  & \et{4.08}{0.07} & \et{1.30}{0.06} & \et{0.63}{0.00}\\
        \bottomrule
    \end{tabular}
    }
\end{table}

\textbf{Quality Indicator Variables.} We remove the motion quality indicator variables from our framework and train the diffusion model using only motion inpainting objectives, while keeping the same network architecture. The model's performance is then evaluated using ground-truth quality labels to identify corrupted frames. As shown in \Cref{tab:appendix_ablation}, the StableMotion model trained without quality indicator variables (w/o \qualvar) underperforms the complete base model. This performance gap highlights the importance of quality indicator variables in enabling the training of an effective motion cleanup model capable of generating high-quality motions.

\textbf{Rotary Positional Embeddings.} We replace the rotary positional embeddings (RoPE) with absolute positional embeddings, which are commonly used in prior works~\citep{tevet2023human,chen2023executing,cohan2024flexible,peebles2023scalable}, while keeping all other settings unchanged, to train a StableMotion model without RoPE. As shown in \Cref{tab:appendix_ablation}, RoPE generally improves artifact cleanup performance across most metrics, particularly for addressing foot skating. However, during training, we observe that RoPE introduces a slight computational overhead compared to absolute positional embeddings, as it requires matrix multiplication in every attention layer. Since our motion cleanup task is performed in an offline setting, this minor latency is acceptable for achieving improved, relatively artifact-free performance.

\subsection{Ablations on Inference Techniques}
\label{subsec:inference_ablation}
We conduct ablation experiments to evaluate the impact of our test-time techniques: \textit{Adaptive Cleanup} and \textit{Quality-Aware Ensemble}. As shown in \Cref{tab:inference_ablation}, the adaptive cleanup technique better preserves the original motion content after cleanup, which is crucial for motion cleanup tasks. The quality-aware ensemble further improves motion quality by leveraging the generative-discriminate model’s motion evaluation capability as a self-evaluation mechanism. Combining both techniques yields complementary benefits, leading to more stable and consistent results. The effectiveness of the quality-aware ensemble is also evident in our qualitative examples provided in the supplementary material.

\begin{table}[t]
    \caption{Ablation experiments on Adaptive Cleanup and Quality-Aware Ensemble.}
    \label{tab:inference_ablation}
    \centering
    \resizebox{\linewidth}{!}{
    \begin{tabular}{l|c|c|c|c|c|c}
        \toprule
        \textbf{Method} & \textbf{FS Dist} $\downarrow$ & \textbf{GP Dist} $\downarrow$ & \textbf{Jitter} $\rightarrow$ & \textbf{GMPJPE} $\downarrow$ & \textbf{M2M Score} $\uparrow$ & \textbf{M2M R@3} $\uparrow$\\
        \midrule
        Clean GT & 3.85 & 4.10 & 1.26 & 0.00 & 1.00 & 100.00 \\
        Input & 6.42 & 4.77 & 7.32 & 1.39 & 0.95 & 88.77 \\
        \midrule
        StableMotion (Base) & \underline{3.44} & \underline{3.88} & 1.14 & 3.59 & 0.92 & 83.04 \\
        + Adaptive Cleanup & 3.84 & 4.36 & \underline{1.20} & \textbf{2.21} & \underline{0.94} & \textbf{85.74} \\
        + Qual Ensemble & \textbf{3.31} & \textbf{3.85} & 1.10 & 3.62 & 0.92 & 82.40 \\
        + Both & 3.71 & 4.12 & \textbf{1.20} & \underline{2.28} & \textbf{0.94} & \underline{85.40} \\
        \bottomrule
    \end{tabular}
    }
\end{table}

\section{Artifacts Synthesis Functions for BrokenAMASS}
\label{sec:appendix_syn}
We describe the artifacts synthesis functions below, with a snippet of the code in Listing~\ref{lst:motion_artifacts}.
We design each atomic motion artifact synthesis function, such as jittering, foot sliding, over-smoothing, and drifting, as an independent motion corruption module, allowing different types of artifact to overlap and create more complex and intractable motion artifacts. 
\begin{itemize}
    \item \textbf{Jittering:} 
    Introduces noise to selected joint positions by adding Gaussian noise with a randomized scale. Specific joints (e.g., all, lower body, left leg, right leg) are selected based on probability. Optionally, a Gaussian smoothing operation is applied to simulate over-smoothing artifacts.
    \item \textbf{Foot Sliding:}
    Alters the root translation in the XY-plane by randomly scaling the actual velocity. This simulates unrealistic foot sliding effects, where the root motion is adjusted to create sliding behavior.
    \item \textbf{Over Smoothing:}
    Applies Gaussian smoothing over the entire motion or specific intervals, which reduces high-frequency motion details and results in unnaturally smooth transitions between frames.
    \item \textbf{Drifting:}
    Gradually shifts the root position over time by introducing a random drift velocity and direction. This creates an unrealistic drifting effect where the motion deviates from its intended trajectory.
\end{itemize}
We randomly select a corruption function and a random interval to corrupt AMASS ~\citep{mahmood2019amass}, leading to an unpaired, corrupted motion dataset, BrokenAMASS, which simulates real-world raw mocap dataset that contains artifacts without paired clean motions.
\begin{lstlisting}[language=Python, caption={Function to synthesis motion artifacts.}, label={lst:motion_artifacts}]
def motion_artifacts(poses, trans, mode='train'):
    """
    Synthesis motion artifacts in joint rotation and translation data.
    
    Args:
        poses (numpy.array): joint rotation of shape (T, J, 4), where T is the number of frames, 
                             J is the number of joints, and 4 represents the quaternion components.
        trans (numpy.array): Translations of shape (T, 3), where T is the number of frames 
                             and 3 represents x, y, z coordinates.
        mode (str): Either 'train' or 'test', determines how augmentations are applied.
    
    Returns:
        poses (numpy.array): Modified poses with simulated artifacts.
        trans (numpy.array): Modified translations with simulated artifacts.
        det_mask (numpy.array): Binary mask indicating artifact locations (1 where artifacts occur).
    """
    # Extract dimensions and motion length
    N, J, D = poses.shape
    mlen = len(poses)

    # Define artifact types
    candidates = ['jittering', 'foot sliding', 'over smooth', 'drifting']
    aug_types = random.sample(candidates, random.randint(1, len(candidates)))  # Randomly pick artifacts
    det_mask = np.zeros_like(trans[:, 0])  # Initialize detection mask for artifacts

    # Skip if motion length is too short
    if mlen < 15:
        return poses, trans, det_mask

    # Apply selected artifact types
    for aug_type in aug_types:
        # Determine artifact length and interval
        if mode == 'train':
            aug_length = min(mlen-2, int(random.randint(5, min(50, mlen-2)) * P))
        else:
            aug_length = random.randint(min(20, int(0.8 * mlen)), min(40, int(0.8 * mlen)))
        aug_interval = random.randint(1, mlen - aug_length)
        det_mask[aug_interval-1: aug_interval + aug_length+1] = 1  # Mark affected frames in the mask

        # Add jittering artifacts
        if aug_type == 'jittering': 
            base_scale = 0.5  # Maximum noise scale
            gaussian_noise_std = 0.1  # Standard deviation for Gaussian noise
            rd = np.random.random() * 0.5 + 0.5  # Random noise scaling factor
            noise_term = np.clip(np.random.randn(N, J, D) * gaussian_noise_std * rd, -base_scale, base_scale)

            # Randomly select joints for applying jitter
            joints_select_id = np.random.choice(np.array([0, 1, 2, 3]), p=np.array([0.4, 0.3, 0.15, 0.15]))
            if joints_select_id == 0:
                joints_selected = random.sample(list(range(J)), random.randint(1, J))  # Random subset of joints
            elif joints_select_id == 1:
                joints_selected = HML_LOWER_BODY_JOINTS[random.choice((0, 1, 3, 5, 7)):]  # Lower body joints
            elif joints_select_id == 2:
                joints_selected = HML_LEFT_LEG_JOINTS[random.choice(range(len(HML_LEFT_LEG_JOINTS))):]  # Left leg joints
            elif joints_select_id == 3:
                joints_selected = HML_RIGHT_LEG_JOINTS[random.choice(range(len(HML_RIGHT_LEG_JOINTS))):]  # Right leg joints
            
            # Apply jittering noise to the selected joints
            poses[aug_interval: aug_interval + aug_length][:, joints_selected] += noise_term[aug_interval: aug_interval + aug_length][:, joints_selected]
            
            # Optional: Add over-smoothing for jittered regions
            if np.random.random() < 0.25:
                radius = round(6 * (np.random.random() * 2 + 2))  # Determine smoothing radius
                poses[aug_interval: aug_interval+aug_length][:, joints_selected] = gaussian_filter1d(
                    poses[:, joints_selected], 
                    sigma=4, 
                    axis=0, 
                    radius=radius, 
                    mode='nearest'
                )[aug_interval: aug_interval + aug_length]

        # Add over-smoothing artifacts
        elif aug_type == 'over smooth':
            radius = round(6 * (np.random.random() * 2 + 2))
            poses[aug_interval: aug_interval+aug_length] = gaussian_filter1d(
                poses, 
                sigma=4, 
                axis=0, 
                radius=radius, 
                mode='nearest'
            )[aug_interval: aug_interval + aug_length]

        # Add foot sliding artifacts
        elif aug_type == 'foot sliding':
            root_disp = trans[:, :2].copy()  # Root displacement in x, y
            root_origin = trans.copy()  # Original root position
            root_origin[:, :2] -= root_disp
            root_vel = root_disp[1:] - root_disp[:-1]  # Root velocity
            root_vel = np.concatenate((np.zeros_like(root_vel[[0]]), root_vel), axis=0)

            scale = 0.1  # Sliding scale
            trans_out = np.zeros_like(trans)
            temp_len = len(trans_out)
            diag_vec = np.ones((temp_len,))
            diag_vec[aug_interval: aug_interval+aug_length] += scale * np.random.random((aug_length,))
            disp_matrix = np.triu(np.broadcast_to(diag_vec[:, None], (temp_len, temp_len)))
            accum_pos_disp = einops.einsum(root_vel, disp_matrix, "n c, n t -> t c")  # Accumulated displacement
            trans_out[:, :2] += accum_pos_disp
            trans_out[:, 2] += trans[:, 2]  # Maintain z-axis
            trans = trans_out

        # Add drifting artifacts
        elif aug_type == 'drifting':
            root_drift_dir = np.random.randn(1, 2) + np.random.randn(aug_length, 2) * 0.1  # Drift direction
            root_drift_vel = np.random.random((aug_length, 1)) * 0.025  # Drift velocity
            root_drift_dir /= np.linalg.norm(root_drift_dir, keepdims=True)  # Normalize direction
            root_drift_vel = root_drift_vel * root_drift_dir
            root_drift_dist = np.cumsum(root_drift_vel, axis=0)  # Drift distance
           trans[aug_interval: aug_interval + aug_length][:, :2] += root_drift_dist  # Apply drift
            trans[aug_interval + aug_length:][:, :2] += root_drift_dist[-1:]  # Apply residual drift

        # Raise error for unknown artifact type
        else:
            raise NotImplementedError
    
    return poses, trans, det_mask
\end{lstlisting}

\section{Additional Generalization Experiements}
\label{sec:appendix_application}

\begin{figure}[t]
    \centering
    \includegraphics[width=\linewidth]{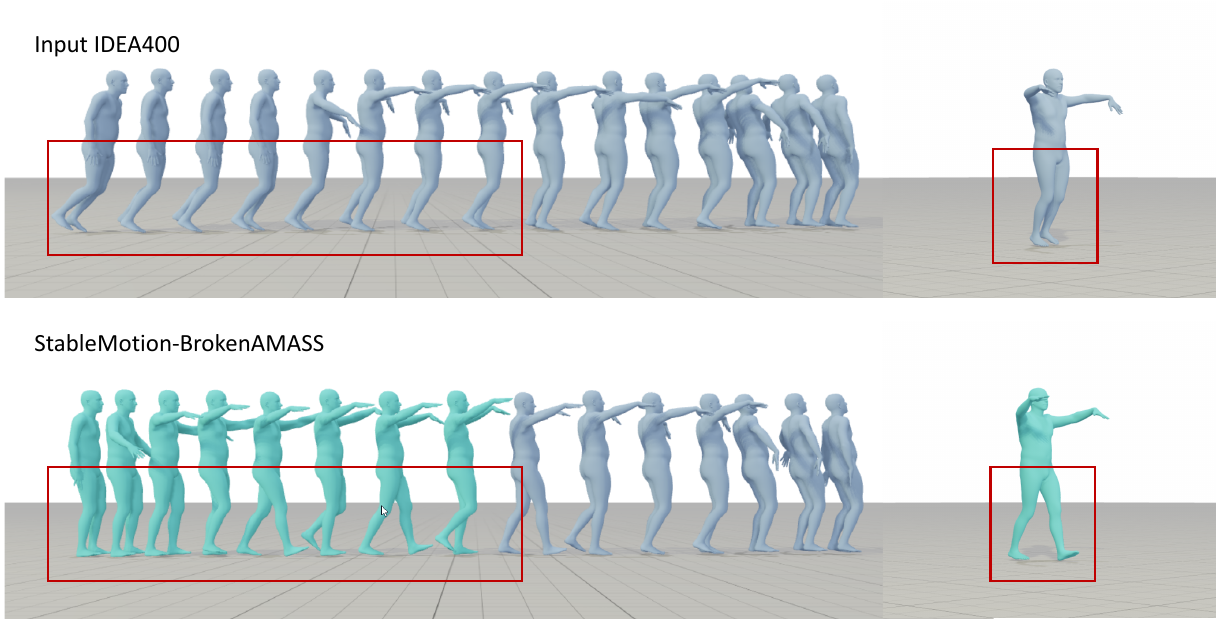}
    \caption{
    Out-of-domain motion cleanup results on the vision-based mocap dataset IDEA400~\citep{lin2024motion}. The trained model can generalize well on out-of-domain data, automatically resolves frozen foot movements, inpainting natural lower-body locomotion while maintaining the zombie-style arm motion.
    }
    \label{fig:idea400}
\end{figure}

To additionally test the generalization ability of our inpainting-based cleanup model, we applied the model trained from BrokenAMASS to new real-world motion data, which are recorded using vision-based motion capture and physics-based character simulations.
First, we apply the trained model to clean up motions from IDEA400~\citep{lin2024motion}, a large-scale vision-based mocap dataset containing 12,042 motion clips. As shown in \Cref{tab:zeroshot}, the trained model effectively removes motion artifacts, particularly jittery motions, while preserving both global and semantic information. It also improves locomotion clips by generating more natural lower-body movements, as illustrated in \Cref{fig:idea400}.

\begin{figure}[t]
    \centering
    \includegraphics[width=\linewidth]{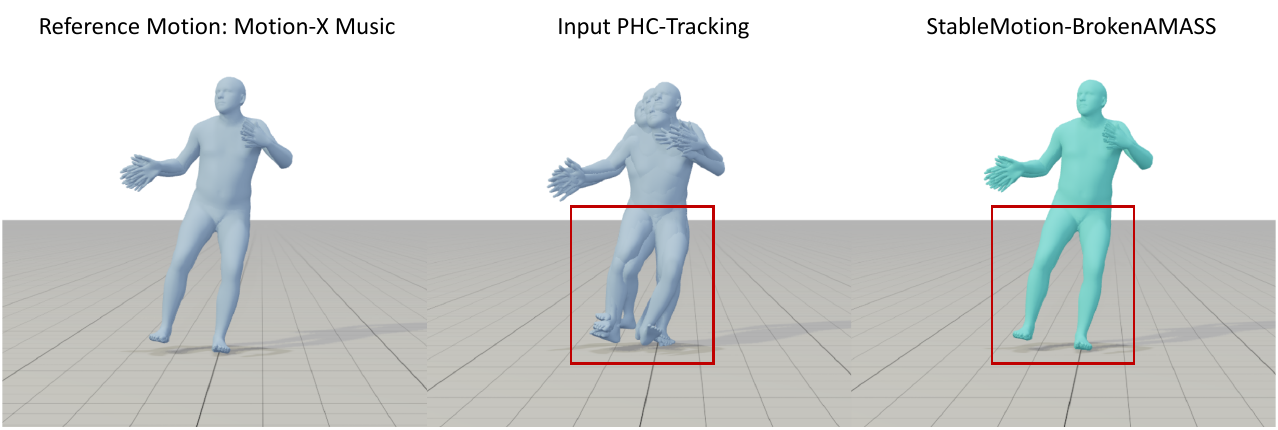}
    \caption{
    While physics-based motion trackers like PHC~\citep{luo2023perpetual} are effective for fixing some physically implausible artifacts, they can also introduce new artifacts, such as jittery foot movements. Our method can also be effective for mitigating artifacts from physically simulated characters.
    }
    \label{fig:music}
\end{figure}

Next, we conduct additional generalization experiments on motions recorded from physics-based character simulations. We use a state-of-the-art physics-based motion tracking model, PHC~\citep{luo2023perpetual}, to track motions from the Motion-X Music subset~\citep{lin2024motion}. The simulated motions are recorded in a motion dataset, PHC-Tracking, containing 2,211 motion clips. The trained model is then applied to clean up PHC-Tracking, and results are reported in \Cref{tab:zeroshot}. While the physics-based tracker can mitigate some artifacts, such as floating, it can also introduce new artifacts, including jittery movements, as illustrated in \Cref{fig:music}. It effectively addresses these jittery movements introduced by PHC and significantly reduces foot skating introduced by the simulation.

\begin{table}[t]
    \caption{Out-of-domain motion cleanup results on the vision-based motion capture dataset (IDEA400~\citep{lin2024motion}) and physics-based simulated character motion (PHC-Tracking~\citep{luo2023perpetual}).} 
    \label{tab:zeroshot}
    \centering
    \resizebox{\linewidth}{!}{
    \begin{tabular}{l|c|c|c|c|c}
        \toprule
        \textbf{Method} & \textbf{FS Dist} $\downarrow$ & \textbf{FS Rate (\%)} $\downarrow$  & \textbf{Jitter} $\downarrow$ & \textbf{GMPJPE (cm)} $\downarrow$ & \textbf{M2M Score} $\uparrow$ \\
        \midrule
        Input: IDEA400 & 1.41 & 2.69 & 2.96 & 0.00 & 1.000 \\
        StableMotion-BrokenAMASS (Base) & 1.14 & 1.30 & 1.50 & 3.43 & 0.906 \\
        StableMotion-BrokenAMASS & 1.21 & 2.32 & 1.77 & 1.38 & 0.962 \\
        \bottomrule
        \toprule
        Input: PHC-Tracking & 0.90 & 2.31 & 1.07 & 0.00 & 1.000 \\
        StableMotion-BrokenAMASS & 0.46 & 0.46 & 0.47 & 1.72 & 0.931\\
        \bottomrule
    \end{tabular}
    }
\end{table}

We also test our model directly on the Motion-X Music subset~\citep{lin2024motion}, where motions are estimated from online single-view videos, containing more artifacts than datasets captured using mocap systems. We compare the trained model against the pretrained state-of-the-art physics-based motion tracking model, PHC~\citep{luo2023perpetual}, which performs motion cleanup through imitation in physics-constrained environments. As shown in \Cref{tab:appendix_music}, the model effectively mitigates motion jittering and achieves comparable performance in fixing foot skating compared to the purely physics-based method. Although both models are trained on datasets sourced from AMASS, our model achieves better content preservation performance, demonstrated by lower joint errors and higher semantic matching scores. This advantage is potentially due to the physics-based motion tracking model's tendency to overfit, while the model built with our framework generalizes better in unseen application scenarios.

\begin{table}[t]
    \caption{Out-of-domain motion cleanup results on Motion-X Music~\citep{lin2024motion}, a motion dataset constructed by human pose estimation from online videos. PHC~\citep{luo2023perpetual}, a SOTA physics-based motion tracking model, is applied to cleanup motion through zero-shot imitation in physics-constrained environments.}
    \label{tab:appendix_music}
    \centering
    \resizebox{\linewidth}{!}{
    \begin{tabular}{l|c|c|c|c|c}
        \toprule
        \textbf{Method} & \textbf{FS Dist} $\downarrow$ & \textbf{FS Rate (\%)} $\downarrow$  & \textbf{Jitter} $\downarrow$ & \textbf{GMPJPE (cm)} $\downarrow$ & \textbf{M2M Score} $\uparrow$ \\
        \midrule
        Input: Motion-X Music& 1.53 & 6.26 & 1.60 & 0.00 & 1.000 \\
        \midrule
        PHC & 0.90 & 2.31 & 1.07 & 2.01 & 0.943 \\
        StableMotion-BrokenAMASS & 0.91 & 1.80 & 0.88 & 0.95 & 0.958 \\
        \bottomrule
    \end{tabular}
    }
\end{table}

Additional generalization experiments on IDEA400, PHC-Tracking, and Motion-X Music demonstrate a secondary application—using a pre-trained cleanup model to clean artifacts from NEW datasets. However, the primary application of StableMotion remains domain-specific cleanup, as demonstrated with in SoccerMocap (\Cref{sec:soccermocapexp}).

\clearpage
\end{document}

%% file: math_commands.tex
\usepackage{amsmath,amsfonts,bm}

\def\eqref#1{equation~\ref{#1}}

\def\1{\bm{1}}

\def\rvb{{\mathbf{b}}}

\def\rvf{{\mathbf{f}}}

\def\rvh{{\mathbf{h}}}

\def\rvm{{\mathbf{m}}}

\def\rvt{{\mathbf{t}}}

\def\rvx{{\mathbf{x}}}
\def\rvy{{\mathbf{y}}}

\DeclareMathAlphabet{\mathsfit}{\encodingdefault}{\sfdefault}{m}{sl}
\SetMathAlphabet{\mathsfit}{bold}{\encodingdefault}{\sfdefault}{bx}{n}

\def\gM{{\mathcal{M}}}

\def\sA{{\mathbb{A}}}
\def\sB{{\mathbb{B}}}

